\documentclass[10pt,journal,compsoc]{IEEEtran}

\ifCLASSINFOpdf
\else
\fi
\usepackage[export]{adjustbox}
\usepackage[skip=0.5ex]{subcaption}
\usepackage{hyperref}
\usepackage{graphicx}
\usepackage{times}
\usepackage{amsmath}
\usepackage{amssymb}
\usepackage{url}
\usepackage{hyperref}
\usepackage{makecell}
\usepackage{ctable}
\usepackage{multirow}
\usepackage{stfloats}
\usepackage{array}
\usepackage{gensymb}
\usepackage{balance}
\usepackage{float}
\usepackage{caption}
\usepackage{subcaption}
\usepackage[font=small,labelfont=bf]{caption}
\usepackage[normalem]{ulem}

\newcolumntype{Q}{>{\centering\arraybackslash}m{1cm}}
\newcolumntype{O}{>{\centering\arraybackslash}m{0.8cm}}
\newcolumntype{P}{>{\centering\arraybackslash}m{1.4cm}}
\newcolumntype{L}{>{\centering\arraybackslash}m{1.5cm}}
\newcolumntype{S}{>{\raggedright\arraybackslash}m{3.8cm}}
\newcolumntype{K}{>{\raggedright\arraybackslash}m{2.3cm}}
\newcolumntype{A}{>{\centering\arraybackslash}m{1.5cm}}
\newcolumntype{R}{>{\centering\arraybackslash}m{0.65cm}}
\newcolumntype{V}{>{\centering\arraybackslash}m{1.3cm}}

\begin{document}

\title{Transformers in Small Object Detection: A Benchmark and Survey of State-of-the-Art}


\author{Aref Miri Rekavandi, \textit{Member, IEEE,} Shima Rashidi, Farid Boussaid, Stephen Hoefs, Emre Akbas, and Mohammed Bennamoun, \textit{Senior Member, IEEE}\\

\thanks{Aref Miri Rekavandi and Mohammed Bennamoun are with the Department of Computer Science and Software Engineering, The University of Western Australia (Emails: aref.mirirekavandi@uwa.edu.au, mohammed.bennamoun@uwa.edu.au). Shima Rashidi is an independent researcher (Email: shima.rashidi7@gmail.com). Farid Boussaid is with the  Department of  Electrical,  Electronics and  Computer Engineering, The University of Western Australia (Email: farid.boussaid@uwa.edu.au). Stephen   Hoefs is a discipline leader at Defence Science and Technology Group, Australia (Email: stephen.hoefs@defence.gov.au). Emre Akbas is with the Department of Computer Engineering, Middle East Technical University, Turkey. (Email:  emre@ceng.metu.edu.tr).}}

\markboth{}%
{Miri \MakeLowercase{\textit{et al.}}: }

\date{}

\IEEEtitleabstractindextext{%

\begin{abstract}
Transformers have rapidly gained popularity in computer vision, especially in the field of object recognition and detection. Upon examining the outcomes of state-of-the-art object detection methods, we noticed that transformers consistently outperformed well-established CNN-based detectors in almost every video or image dataset. While transformer-based approaches remain at the forefront of small object detection (SOD) techniques, this paper aims to explore the performance benefits offered by such extensive networks and identify potential reasons for their SOD superiority. Small objects have been identified as one of the most challenging object types in detection frameworks due to their low visibility. We aim to investigate potential strategies that could enhance transformers' performance in SOD. This survey presents a taxonomy of over 60 research studies on developed transformers for the task of SOD, spanning the years 2020 to 2023. These studies encompass a variety of detection applications, including small object detection in generic images, aerial images, medical images, active millimeter images, underwater images, and videos. We also compile and present a list of 12 large-scale datasets suitable for SOD that were overlooked in previous studies and compare the performance of the reviewed studies using popular metrics such as mean Average Precision (mAP), Frames Per Second (FPS), number of parameters, and more. Researchers can keep track of newer studies on our web page, which is available at: \url{https://github.com/arekavandi/Transformer-SOD}.  
\end{abstract}

\begin{IEEEkeywords}
Object recognition, small object detection, vision transformers, object localization, deep learning, attention, MS COCO dataset.
\end{IEEEkeywords}}

\maketitle

\IEEEpeerreviewmaketitle

\section{Introduction}\label{Introduction}
\begin{figure}
\begin{center}
   \includegraphics[width=0.46\linewidth]{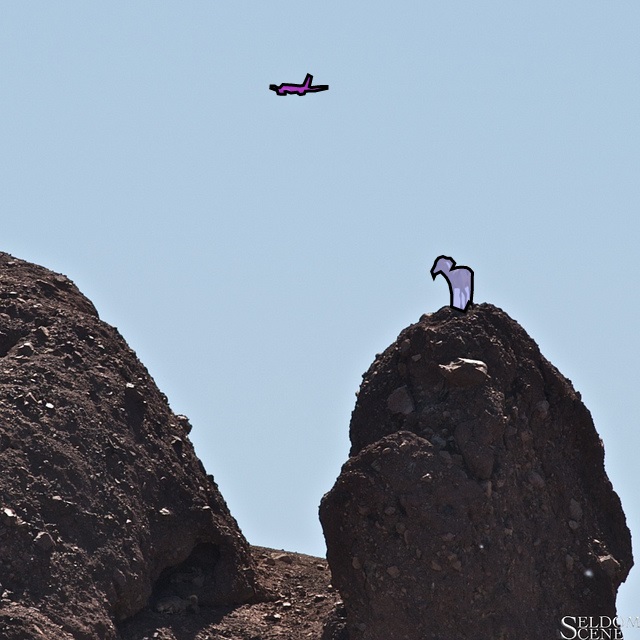}
   \includegraphics[width=0.305\linewidth]{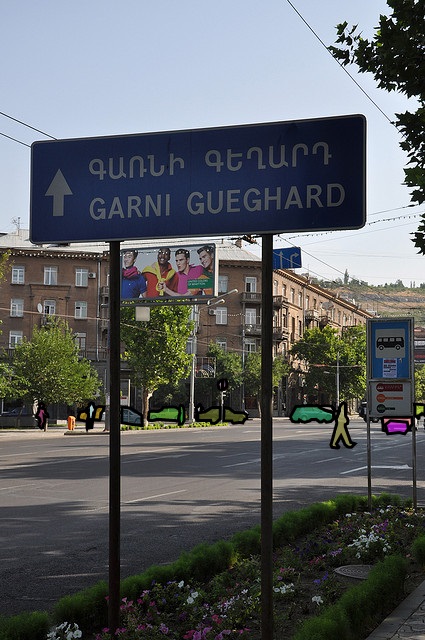}
      \includegraphics[width=0.23\linewidth]{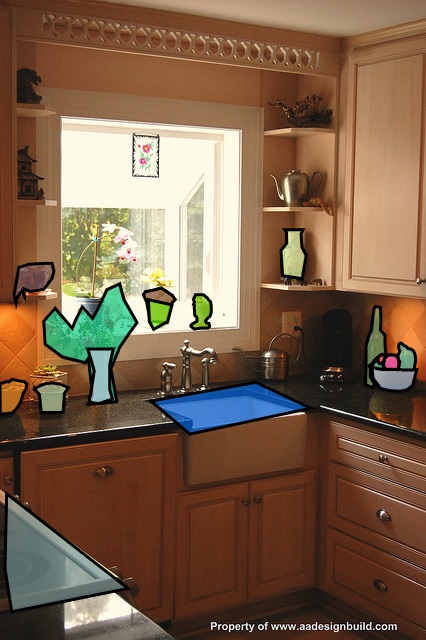}
   \includegraphics[width=0.55\linewidth]{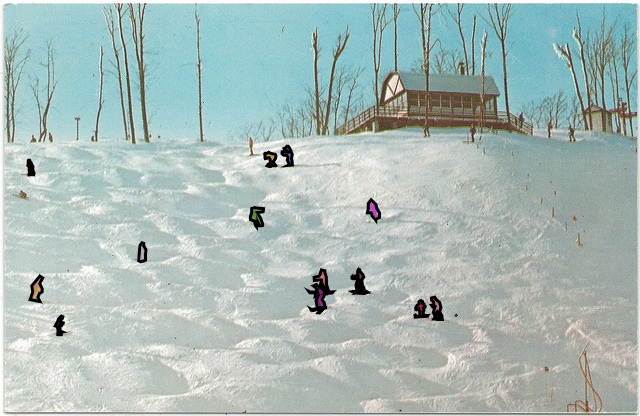}
\end{center}
   \caption{Examples of small size objects from MS COCO dataset \cite{lin2014microsoft}. The objects are highlighted with color segments.}
\label{fig1}
\end{figure}

\IEEEPARstart{S}{mall} Object Detection (SOD) has been recognized as a significant challenge for State-Of-The-Art (SOTA) object detection methods \cite{liu2021survey}. The term ``small object'' refers to objects that occupy a small fraction of the input image. For example, in the widely used MS COCO dataset \cite{lin2014microsoft}, it defines objects whose bounding box is $32 \times 32$ pixels or less, in a typical $480 \times 640$ image (Figure \ref{fig1}). Other datasets have their own definitions, e.g. objects that occupy $10\%$ of the image.  Small objects are often missed or detected with incorrectly localized bounding boxes, and sometimes with incorrect labels. 
The main reason for the deficient localization in SOD stems from the limited information provided in the input image or video frame, compounded by the subsequent spatial degradation experienced as they pass through multiple layers in deep networks. Since small objects frequently appear in various application domains, such as pedestrian detection \cite{wu2020self}, medical image analysis \cite{rashidi2020optimal}, face recognition \cite{cho2018face}, traffic sign detection \cite{liu2019mr}, traffic light detection \cite{yudin2018usage}, ship detection \cite{varga2021tackling}, Synthetic Aperture Radar (SAR)-based object detection \cite{rekavandi2021robust}, it is worth examining the performance of modern deep learning SOD techniques. In this paper, we compare transformer-based detectors with Convolutional Neural Networks (CNNs) based detectors in terms of their small object detection performance. In the case of outperforming CNNs with a clear margin, we then attempt to uncover the reasons behind the transformer’s strong performance.  One immediate explanation could be that transformers model the interactions between pairwise locations in the input image. This is effectively a way of encoding the context. And, it is well established that context is a major source of information to detect and recognize small objects both in humans and computational models \cite{contextual_priming}.  However, this might not be the only factor to explain transformers’ success. Specifically, we aim to analyze this success along several dimensions including object representation, fast attention for high-resolution or multiscale
feature maps, fully transformer-based detection, architecture
and block modification, auxiliary techniques, improved feature
representation, and spatio-temporal information.  Furthermore, we point out approaches that could potentially enhance the performance of transformers for SOD. 

In our previous work, we surveyed numerous strategies
employed in deep learning to enhance the performance of small
object detection in optical images and videos up to the year 2022 \cite{rekavandi2022guide}. We showed that beyond the adaptation of newer deep learning structures such as transformers, prevalent approaches include data augmentation, super-resolution, multi-scale feature learning, context learning, attention-based learning, region proposal, loss function regularization, leveraging auxiliary tasks, and spatiotemporal feature aggregation. Additionally, 
we observed that transformers are among the leading methods in localizing
small objects across most datasets. However, given that \cite{rekavandi2022guide} predominantly evaluated over 160 papers focusing
on CNN-based networks, an in-depth exploration of transformer-centric methods was not undertaken. Recognizing the growth and
exploration pace in the field, there is a timely window now to
delve into the current transformer models geared towards small
object detection.

In this paper, our goal is to comprehensively
understand the factors contributing to the impressive performance
of transformers when applied to small object detection and their distinction with strategies used for generic object detection. To lay
the groundwork, we first highlight renowned transformer-based
object detectors for SOD, juxtaposing their advancements against
established CNN-based methodologies.

Since 2017, the field has seen the publication of numerous
review articles. An extensive discussion and listing of these reviews
are presented in our previous survey \cite{rekavandi2022guide}. Another recent survey article \cite{cheng2023towards}  mostly focuses on the CNN-based
techniques, too. The narrative of this current survey stands distinct
from its predecessors. Our focus in this paper narrows down
specifically to transformers — an aspect not explored previously — positioning them as the dominant network architecture for image and video SOD. This entails a unique taxonomy tailored to this innovative architecture, consciously sidelining CNN-based methods. Given the novelty and intricacy of this topic, our review prioritizes works primarily brought forth post-2022. Additionally, we shed light on newer datasets employed for the localization and detection of small objects across a broader spectrum of applications.

The studies examined in this survey primarily presented methods
tailored for small object localization and classification or
indirectly tackled SOD challenges. What drove our analysis were
the detection outcomes specified for small objects in these papers.
However, earlier research that noted SOD outcomes but either
demonstrated subpar performance or overlooked SOD-specific
parameters in their development approach were not considered
for inclusion in this review. In this survey, we assume the reader is already familiar with
generic object detection techniques, their architectures, and relevant
performance measures. If the reader requires foundational
insight into these areas, we refer the reader to our previous work
\cite{rekavandi2022guide}.

The structure of this paper is as follows: Section \ref{background} offers an overview of CNN-based object detectors, transformers, and their components, including the encoder and decoder. This section also touches upon two initial iterations of transformer-based object detectors: DETR and ViT-FRCNN. In Section \ref{Methods}, we present a classification for transformer-based SOD techniques and delve into each category comprehensively. Section \ref{Results} showcases the different datasets used for SOD and evaluates them across a range of applications. In Section \ref{findings},  we analyze and contrast these outcomes with earlier results derived from CNN networks. The paper wraps up with conclusions in Section \ref{conclusion}.

\section{Background} \label{background}
Object detection and in particular SOD, has long relied on CNN-based deep learning models. Several single-stage and two-stage detectors  have emerged over time, such as You Only Look Once (YOLO) variants \cite{redmon2016you,redmon2017yolo9000,redmon2018yolov3,bochkovskiy2020yolov4,jocher2020yolov5,li2022yolov6,wang2022yolov7}, Single Shot multi-box Detector (SSD) \cite{liu2016ssd}, RetinaNet \cite{lin2017focal}, Spatial Pyramid Pooling Network (SPP-Net) \cite{he2015spatial},  Fast R-CNN \cite{girshick2015fast}, Faster R-CNN \cite{ren2015faster}, Region-Based Fully Convolutional Networks (R-FCN) \cite{dai2016r}, Mask R-CNN \cite{he2017mask}, Feature Pyramid Networks  (FPN) \cite{lin2017feature}, cascade R-CNN \cite{cai2018cascade}, and  Libra R-CNN \cite{pang2019libra}. Various strategies have been used in conjunction with these techniques to improve their detection performance for SOD, with multi-scale feature learning being the most commonly used approach.\\
\begin{figure}
\begin{center}
   \includegraphics[width=0.9\linewidth]{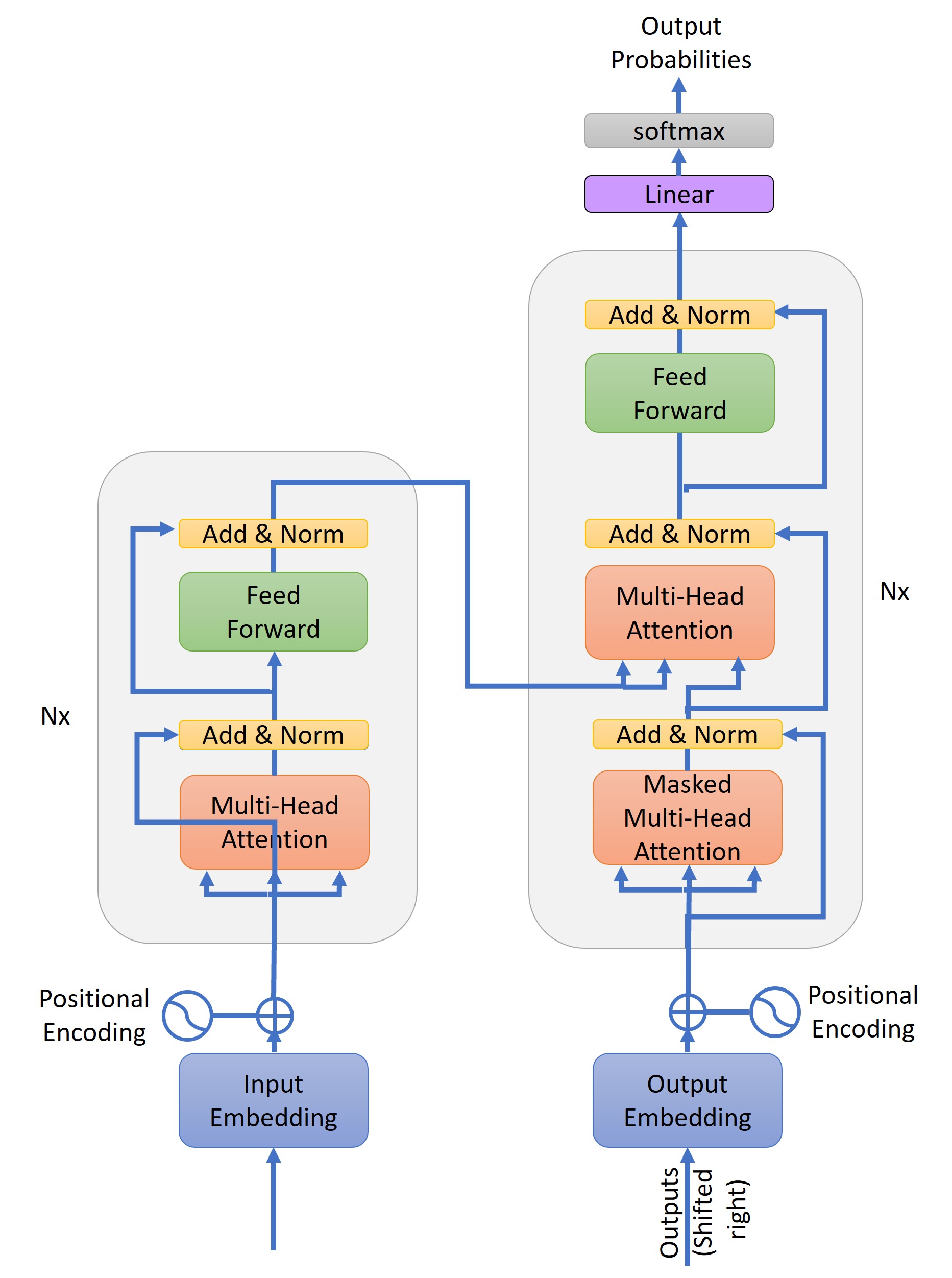}
\end{center}
   \caption{Transformer architecture containing encoder (left module) and decoder (right module) used in sequence to sequence translation (figure from \cite{vaswani2017attention}).}
\label{fig2}
\end{figure}

\begin{figure*}
\begin{tabular}{c}
    \includegraphics[width=0.95\linewidth]{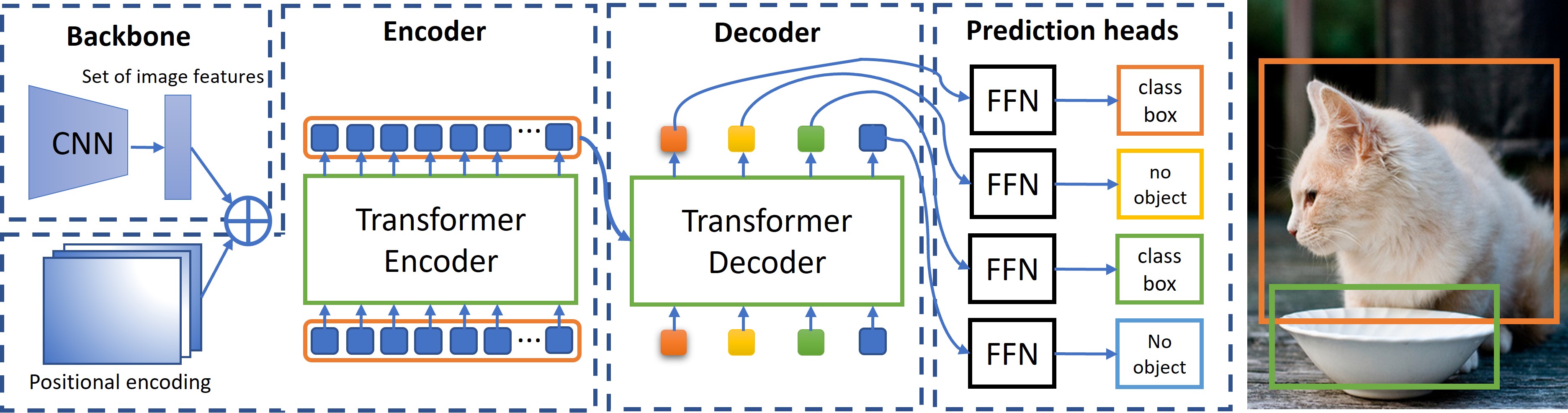}  \\
    \hline
     \includegraphics[width=0.95\linewidth]{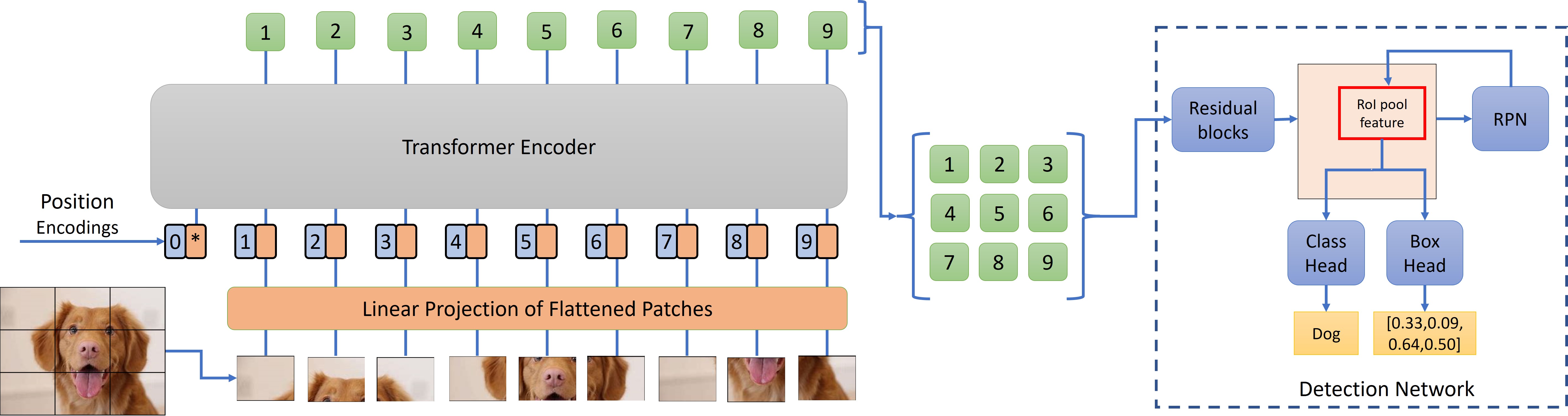}
\end{tabular}  
   \caption{Top: DETR (figure from \cite{carion2020end}). Bottom: ViT-FRCNN (figure from \cite{beal2020toward}).}
\label{back}
\end{figure*}
\begin{table*}[t!]
\caption{A list of terminologies used in this paper with their meanings.}
  \centering
 \begin{adjustbox}{scale=0.99,angle=0}
    \begin{tabular}{ll}
    \hline
    Full Term & Description \\
    \hline
      Encoder & Encoder in transformers consists of multiple layers of self-attention modules and feed-forward neural networks to extract local  \\
      & and global semantic information from the input data.\\
      Decoder & Decoder module is responsible to generate the output (either sequence or independent) based on the concept of self and cross\\
      & attention applied to the object queries and encoder's output.\\
      Token &  Token refers to the most basic unit of data input into the transformers. It can be image pixels, patches, or video clips.\\
      Multi-Head Attention & Multi-Head Attention is a mechanism in transformers that enhances the learning capacity and representational power of \\
      &  self-attention. It divides the input into multiple subspaces and performs attention computations independently on each subspace, \\
      & known as attention heads.\\
      Spatial Attention & Spatial attention in transformers refers to a type of attention mechanism that attends to the spatial positions of tokens within a \\
      & sequence. It allows the model to focus on the relative positions of tokens and capture spatial relationships.\\
      Channel Attention & Channel attention in transformers refers to an attention mechanism that operates across different channels or feature dimensions of \\
      & the input. It allows the model to dynamically adjust the importance of different channels, enhancing the representation and modeling \\
      & of channel-specific information in tasks\\
      Object Query & It refers to a learned vector representation that is used to query and attend to specific objects or entities within a scene. \\
      Positional Embedding & It refers to a learned representation that encodes the positional information of tokens in an input sequence, enabling the model to \\
      & capture sequential dependencies.\\
    \bottomrule
    \end{tabular}%
    \end{adjustbox}
  \label{terms}%
\end{table*}%
The transformer model was first introduced in \cite{vaswani2017attention} as a novel technique for machine translation. This model aimed to advance beyond traditional recurrent networks and CNNs by introducing a new network architecture solely based on attention mechanisms, thereby eliminating the need for recurrence and convolutions. The Transformer model consists of two main modules: the encoder and the decoder. Figure \ref{fig2} provides a visual representation of the processing blocks within each module. The description of terminologies commonly used in Transformers for computer vision is provided in Table \ref{terms} for readers who are not familiar with the topic.
Within the context of SOD, the encoder module ingests input tokens, which can refer to image patches or video clips, and employs various feature embedding approaches, such as utilizing pre-trained CNNs to extract suitable representations. The positional encoding block embeds positional information into the feature representations of each token. Positional encoding has demonstrated significant performance improvements in various applications. The encoded representations are then passed through a Multi-Head Attention block, which is parameterized with three main matrices, namely $\textbf{W}_q \in \mathbf{R}^{d_q \times d}$, $\textbf{W}_k\in \mathbf{R}^{d_k \times d}$, and $\textbf{W}_v\in \mathbf{R}^{d_v \times d}$ to obtain query, key and value vectors, shown by $\textbf{q}$, $\textbf{k}$, $\textbf{v}$, respectively. In other words,
\begin{equation}
    \textbf{q}_i=\textbf{W}_q\textbf{x}_i,\quad \textbf{k}_i=\textbf{W}_k\textbf{x}_i, \quad \textbf{v}_i=\textbf{W}_v\textbf{x}_i,\quad i=1,\cdots,T,
\end{equation}
where $T$ is the total number of tokens and each token is denoted by $\textbf{x}$. The output of the Multi-Head Attention block is given by
\begin{equation}
    \text{MH Attention}(\textbf{Q}, \textbf{K}, \textbf{V})=\text{Concat}(\text{head}_1,\cdots,\text{head}_h)\textbf{W}^O.
\end{equation}
where $\textbf{W}^O \in \mathbf{R}^{hd_v \times d}$, $d_k=d_q$, and
\begin{equation}\label{this}
    \text{head}_h=\text{Attention}(\textbf{Q}_h, \textbf{K}_h, \textbf{V}_h)=\text{Softmax}(\frac{\textbf{K}_h^\top\textbf{Q}_h}{\sqrt{d_k}})\textbf{V}_h^\top.
\end{equation}
Finally, the results obtained from the previous steps are combined with a skip connection and a normalization block. These vectors are then individually passed through a fully connected layer, applying an activation function to introduce non-linearity into the network. The parameters of this block are shared across all vectors. This process is repeated for a total of $N$ times, corresponding to the number of layers in the deep network.
In the decoder module, a similar process is applied using the vectors generated in the encoder, while also consuming the previously generated predictions/outputs as additional input. Ultimately, the output probabilities for the possible output classes are computed. Attention is achieved through the dot product operation between the key and query matrices, Eq. (\ref{this}), which computes weights for the linear combination of the matrix $\textbf{V}$.

An alternative representation for the transformer is also provided in
\begin{equation}
    \text{MH Attention}^i=\sum_{h}\textbf{W}^O_h\big[\sum_{k=1}^{T}A_{hik}\textbf{W}_v\textbf{x}_k\big], i=1,\cdots,T,
\end{equation}
where $\textbf{W}^O_h$ is a submatrix of $\textbf{W}^O$ that corresponds to $h-$th head, and $A_{hik}$ is the attention weight in $h-$th head which is the element in $i-$th row (corresponds to $i-$th query) and $k-$th column  (corresponds to $k-$th key) of the matrix: $\text{Softmax}(\frac{\textbf{K}_h^\top\textbf{Q}_h}{\sqrt{d_k}})$.

Dosovitskiy \textit{et al.} were the first to utilize the architecture of transformers in computer vision tasks, including image recognition \cite{dosovitskiy2020image}. The remarkable performance exhibited by transformers in various vision tasks has paved the way for their application in the domain of object detection research. Two pioneering works in this area are the DEtection TRansformer (DETR) \cite{carion2020end} (Figure \ref{back}, Top) and ViT-FRCNN \cite{beal2020toward} (Figure \ref{back}, Bottom).

DETR aimed to reduce the reliance on CNN-based techniques during post-processing by employing a set-based global loss. This particular loss function aids in the collapse of near-duplicate predictions through bipartite matching, ensuring each prediction is uniquely paired with its matching ground truth bounding boxes. As an end-to-end model, DETR benefits from global computation and perfect memory, making it suitable for handling long sequences generated from videos/images. The bipartite matching loss utilized in DETR is defined as follows:
\begin{figure*}
\begin{center}
   \includegraphics[width=1\linewidth]{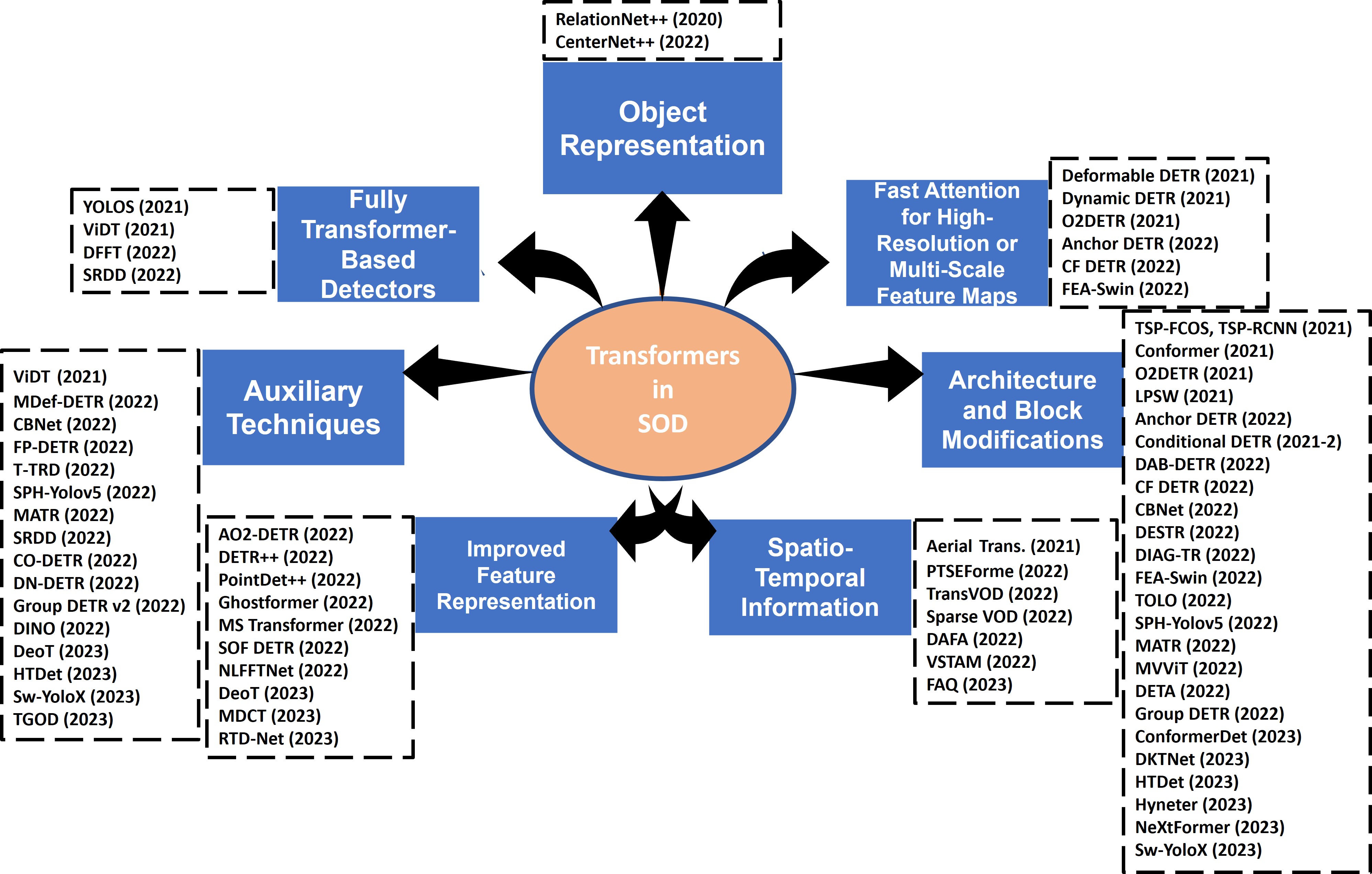}
\end{center}
   \caption{Taxonomy of small object detection using transformers and popular object detection methods in each category.}
\label{fig3}
\end{figure*}
\begin{equation}
    \hat{s}=\arg\min_{s \in \mathcal{S}}\sum_{i}^{N}\mathcal{L}
    _{match}(y_i,\hat{y}_{s(i)}),
\end{equation}
where $\mathcal{L}_{match}(y_i,\hat{y}_{s(i)})$ measures the pair-wise matching cost between ground truth box $y_i$ with size $N$ and the prediction with index $s(i)$ where $s$ is a specific order of predicted bounding boxes. In this formulation, $N$ is the largest possible number of objects within an image. In the case of fewer objects in predictions and ground-truth, $y$ and $\hat{y}$ will be padded with $\emptyset$ (indicating no object). Consequently, this loss function considers all possible matching policies between predictions and ground truth, selecting the one that yields the minimum loss value. The optimal pairing can be efficiently computed using the Hungarian algorithm, as demonstrated in \cite{stewart2016end}. DETR used a CNN backbone to extract compact feature representations and an encoder-decoder transformer with a feed-forward network to produce the final predictions (see Figure \ref{back}, Top). In contrast, ViT-FRCNN uses the Vision Transformer (ViT) \cite{dosovitskiy2020image} for object detection and demonstrates that pre-trained ViT on large-scale datasets enhances the detection performance through rapid fine-tuning. While ViT-FRCNN, like DETR, incorporates CNN-based networks in its pipeline, specifically in the detection head, it diverges from DETR by using the Transformer (encoder only) to encode visual attributes. Additionally, a conventional Region Proposal Network (RPN) \cite{ren2015faster} is used for generating detections (illustrated in Figure \ref{back}, Bottom).
Both DETR and ViT-FRCNN have shown subpar results in the detection and classification of small objects. ViT-FRCNN even exhibited worse results when increasing the token size of the input image. The best outcomes were achieved when the token size was set to $16 \times 16$, and all intermediate transformer states were concatenated with the final transformed layer. Additionally, both detectors rely on CNNs at different stages, in DETR as the backbone for feature extraction and in ViT-FRCNN for the detection head. To improve the results of small object detection, it is crucial to retain the image patches as small as possible to preserve spatial resolution, which consequently increases the computational costs. To address these limitations and challenges, further research has been conducted, which will be discussed in detail in the following sections.

\begin{figure*}
\begin{center}
   \includegraphics[width=0.9\linewidth]{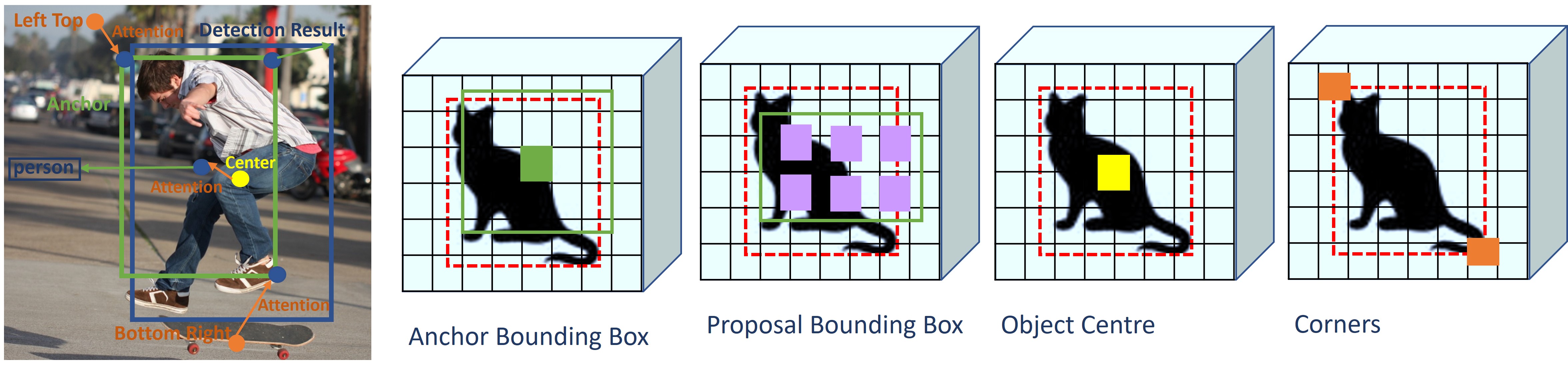}
\end{center}
   \caption{BVR uses different representations. i.e., corner and center points to enhance features for anchor-based detection (left figure). Object representations are shown for another image (cat) where red dashes show the ground truth (figure from \cite{chi2020relationnet++}).}
\label{BVR}
\end{figure*}
\section{Transfomers For Small Object Detection}\label{Methods}

In this section, we discuss transformer-based networks for SOD. A taxonomy of small object detectors is shown in Figure \ref{fig3}. We show that existing detectors based on novel transformers can be analyzed through one or a few of the following perspectives: object representation, fast attention for high-resolution or multi-scale feature maps, fully transformer-based detection, architecture and block modification, auxiliary techniques, improved feature representation, and spatio-temporal information. In the following subsections, each of these categories is discussed in detail separately.

\subsection{Object Representation}
Various object representation techniques have been adopted in object detection techniques. The object of interest can be represented by rectangular boxes \cite{girshick2015fast}, points such as center points \cite{zhou2019objects} and point sets \cite{yang2019reppoints}, probabilistic objects \cite{wang2021normalized}, and keypoints \cite{law2018cornernet}. Each object representation technique has its own strengths and weaknesses, with respect to the need for annotation formats and small object representation. The pursuit of finding the optimal representation technique, while keeping all the strengths of the existing representations, began with RelationNet++ \cite{chi2020relationnet++}. This approach bridges various heterogeneous visual representations and combines their strengths via a module called Bridging Visual Representations (BVR). BVR operates efficiently without disrupting the overall inference process employed by the main representations, leveraging novel techniques of key sampling and shared location embedding. More importantly, BVR relies on an attention module that designates one representation form as the ``master representation'' (or query), while the other representations are designated as ``auxiliary'' representations (or keys). The BVR block is shown in Figure \ref{BVR}, where it enhances the feature representation of the anchor box by seamlessly integrating center and corner points (keys) into the anchor-based (query) object detection methodology. Different object representations are also shown in Figure \ref{BVR}. CenterNet++ \cite{duan2022centernet++} was proposed as a novel bottom-up approach. Instead of estimating all the object's parameters at once, CenterNet++ strategically identifies individual components of the object separately, i.e., top-left, bottom-left, and center keypoints. Then, post-processing methodologies are adopted to cluster points associated with the same objects. This technique has demonstrated a superior recall rate in SOD compared to top-down approaches that estimate entire objects as a whole.

\begin{figure}
\begin{center}
   \includegraphics[width=0.9\linewidth]{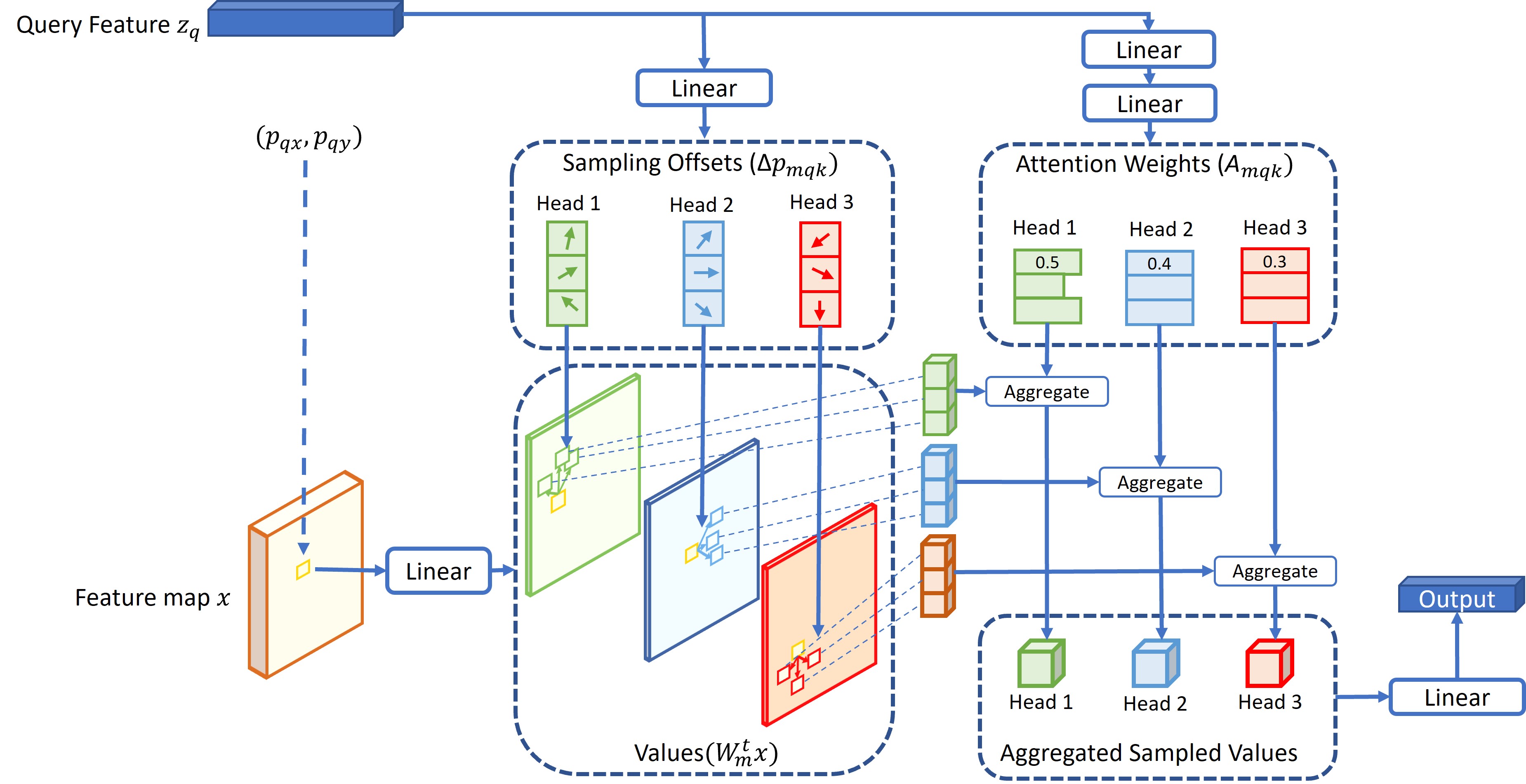}
\end{center}
   \caption{The block diagram for the Deformable attention module. $\textbf{z}_q$ is the content feature of the query, $\textbf{x}$ is the feature map and $\textbf{p}_q$ is the refrence point in 2-D grid. In short, the deformable attention module only attends to a small set of key sampling points around the reference point (different in each head). This significantly reduces the complexity and further improves the convergence (figure from \cite{zhu2021deformable}).}
\label{Deformable}
\end{figure}
\subsection{Fast Attention for High-Resolution or Multi-Scale Feature Maps}
Previous research has shown that maintaining a high resolution of feature maps is a necessary step for maintaining high performance in SOD. Transformers, inherently exhibit a notably higher complexity compared to CNNs due to their quadratic increase in complexity with respect to the number of tokens (e.g., pixel numbers). This complexity emerges from requirement of pairwise correlation computation across all tokens. Consequently, both training and inference times exceed expectations, rendering the detector inapplicable for small object detection in high-resolution images and videos.
\begin{figure*}
\begin{center}
   \includegraphics[width=0.9\linewidth]{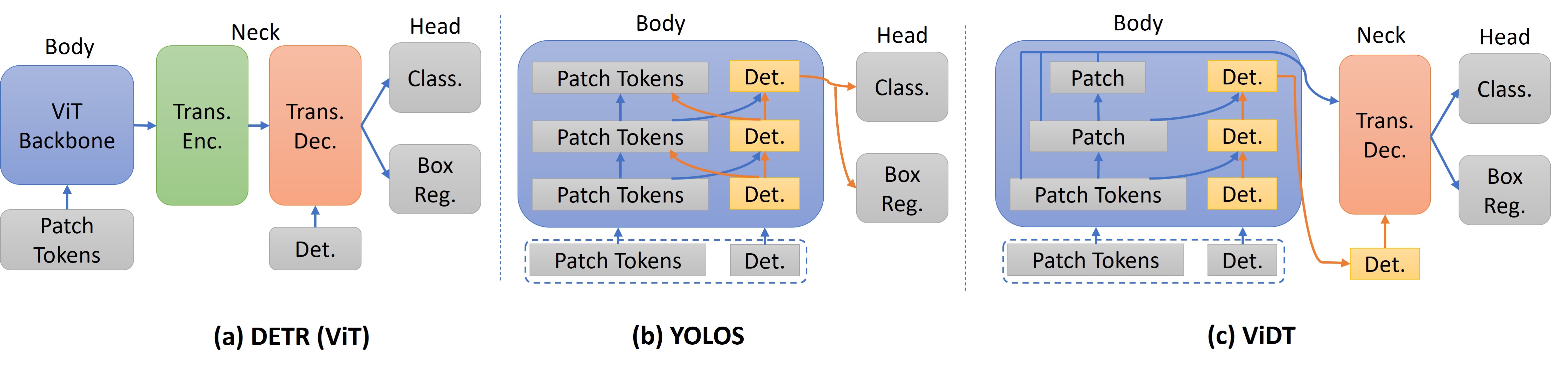}
\end{center}
   \caption{ViDT (c) mixes DETR (with ViT backbone or other fully transformer-based backbones) (a) with YOLOS architecture (b) in a multi-scale feature learning pipeline to achieve SOTA results (figure from \cite{song2022vidt}).}
\label{ViDT}
\end{figure*}
In their work on Deformable DETR, Zhu \textit{et al.} \cite{zhu2021deformable} addressed this issue that had been observed in DETR for the first time. They proposed attending to only a small set of key sampling points around a reference, significantly reducing the complexity. By adopting this strategy, they effectively preserved spatial resolution through the use of multi-scale deformable attention modules. Remarkably, this method eliminated the necessity for feature pyramid networks, thereby greatly enhancing the detection and recognition of small objects. The $i-$th output of a multi-head attention module in Deformable attention is given by:
\begin{equation}
    \text{MH Attention}^i=\sum_{h}\textbf{W}^O_h\big[\sum_{k=1}^{K}A_{hik}\textbf{W}_v\textbf{x}_k(\textbf{p}_i+\Delta\textbf{p}_{hik})\big],
\end{equation}
where $i=1,\cdots,T$ and $\textbf{p}_i$ is the reference point of the query and $\Delta\textbf{p}_{hik}$ is the sampling offset (in 2D) in $h-$th head with K samplings (K$<<$T=HW). Figure \ref{Deformable} illustrates the computation process within its multi-head attention module.
Deformable DETR benefits from both its encoder and decoder modules, with the complexity order within the encoder being $\mathcal{O}(HWC^2)$ where $H$ and $W$ are the height and width of input feature map and $C$ is the number of channels. In contrast for the DETR encoder, the order of complexity is $\mathcal{O}(H^2W^2C)$, displaying a quadratic increase as $H$ and $W$ increase in size. Deformable attention has played a prominent role in various other detectors, e.g., in T-TRD \cite{li2022transformer}. Subsequently, Dynamic DETR was proposed in \cite{dai2021dynamic}, featuring a dynamic encoder and a dynamic decoder that harness feature pyramids from low to high-resolution representations, resulting in efficient coarse-to-fine object detection and faster convergence. The dynamic encoder can be viewed as a sequentially decomposed approximation of full self-attention, dynamically adjusting  attention mechanisms based on scale, spatial importance, and representation. Both Deformable DETR and Dynamic DETR make use of deformable convolution for feature extraction. In a distinct approach, O$^2$DETR \cite{ma2021oriented} demonstrated that the global reasoning offered by a self-attention module is actually not essential for aerial images, where objects are usually densely packed in the same image area. Hence, replacing attention modules with local convolutions coupled with the integration of multi-scale feature maps, was proven to improve the detection performance in the context of oriented object detection. The authors in \cite{wang2022anchor} proposed the concept of Row-Column Decoupled
Attention (RCDA), decomposing the 2D attention of key features into two simpler forms: 1D row-wise and column-wise attentions. In the case of CF-DETR \cite{cao2022cf}, an alternative approach to FPN was proposed whereby C5 features were replaced with encoder features at level 5 (E5), resulting in improved object presentation. This innovation was named Transformer Enhanced FPN (TEF) module. In another study, Xu~\textit{et al.} \cite{xu2022fea} developed a weighted Bidirectional Feature Pyramid Network (BiFPN) through the integration of skip connection operations with the Swin transformer. This approach effectively preserved information pertinent to small objects.

\subsection{Fully Transformer-Based Detectors}
The advent of transformers and their outstanding performance in many complex tasks in computer vision has gradually motivated researchers to shift from CNN-based or mixed systems to fully transformer-based vision systems. This line of work started with the application of a transformer-only architecture to the image recognition task, known as ViT, proposed in \cite{dosovitskiy2020image}. In \cite{song2022vidt}, ViDT extended the YOLOS model \cite{fang2021you} (the first fully transformer-based detector) to develop the first efficient detector suitable for SOD. In ViDT, the ResNet used in DETR for feature extraction is replaced with various ViT variants, such as Swin Transformer \cite{liu2021swin}, ViTDet \cite{li2022exploring}, and DeiT \cite{touvron2021training}, along with the Reconfigured Attention Module (RAM). The RAM is capable of handling $[\text{PATCH}] \times [\text{PATCH}]$, $[\text{DET}] \times [\text{PATCH}]$, and $[\text{DET}] \times [\text{DET}]$ attentions. These cross and self-attention modules are necessary because, similar to YOLOS, ViDT appends $[\text{DET}]$ and $[\text{PATCH}]$ tokens in the input. ViDT only utilizes a transformer decoder as its neck to exploit multi-scale features generated at each stage of its body step. Figure \ref{ViDT} illustrates the general structure of ViDT and highlights its differences from DETR and YOLOS.

Recognizing that the decoder module is the main source of inefficiency in transformer-based object detection, the Decoder-Free Fully Transformer (DFFT) \cite{chen2022efficient} leverages two encoders: Scale-Aggregated Encoder (SAE) and Task-Aligned Encoder (TAE), to maintain high accuracy. SAE aggregates the multi-scale features (four scales) into a single feature map, while TAE aligns the single feature map for object type and position classification and regression. Multi-scale feature extraction with strong semantics is performed using a Detection-Oriented Transformer (DOT) backbone.

In Sparse RoI-based deformable DETR (SRDD) \cite{zhu2022srdd}, the authors proposed a lightweight transformer with a scoring system to ultimately remove redundant tokens in the encoder. This is achieved using RoI-based detection in an end-to-end learning scheme.

\subsection{Architecture and Block Modifications}
DETR, the first end-to-end object detection method, strugles with extended converge times during training and performs poorly on small objects. Several research works have addressed these issues to improve SOD performance. One notable contribution comes from Sun et al. \cite{sun2021rethinking}, who, drawing inspiration from FCOS \cite{tian2019fcos} (a fully convolutional single-stage detector) and Faster RCNN, proposed two encoder-only DETR variants with feature pyramids called TSP-FCOS and TSP-RCNN. This was accomplished by eliminating cross-attention modules from the decoder. Their findings demonstrated that cross-attention in the decoder and the instability of the Hungarian loss were the main reasons for the late convergence in DETR. This insight led them to discard the decoder and introduce a new bipartite matching technique in these new variants, i.e., TSP-FCOS and TSP-RCNN.

\begin{figure*}
\begin{center}
   \includegraphics[width=0.7\linewidth]{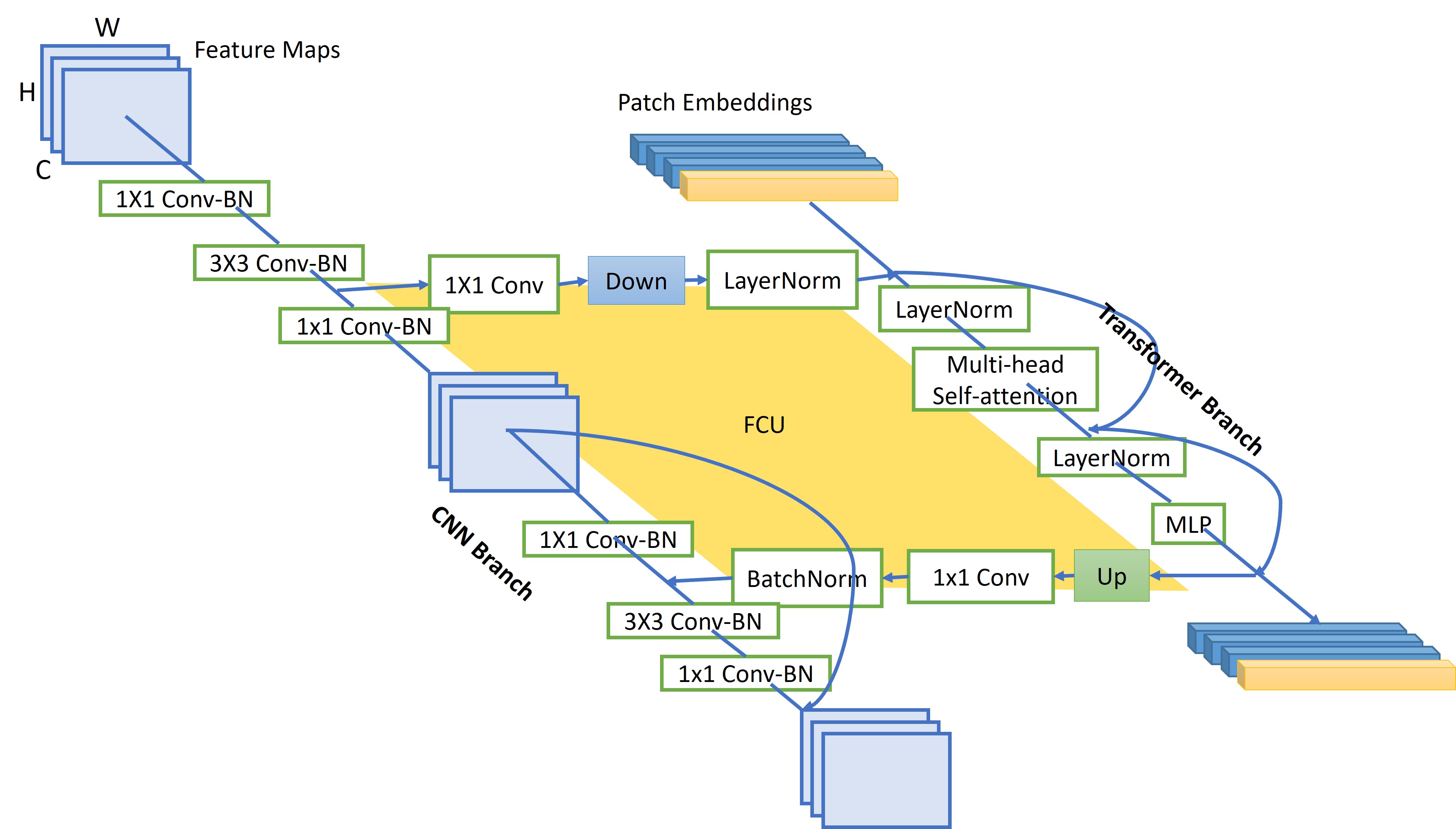}
\end{center}
   \caption{Conformer architecture which leverages both local features provided by CNNs and global features provided by transformers in Feature Coupling Unit (FCU) (figure from \cite{peng2023conformer}).}
\label{conformer}
\end{figure*}

In a combined approach using CNNs and transformers, Peng et al. \cite{peng2021conformer, peng2023conformer} proposed a hybrid network structure called ``Conformer''. This structure fuses the local feature representation provided by CNNs with the global feature representation provided by transformers at varying resolutions (see Figure \ref{conformer}). This was achieved through Feature Coupling Units (FCUs), with experimental results demonstrating its effectiveness compared to ResNet50, ResNet101, DeiT, and other models. A similar hybrid technique combining CNNs and transformers was proposed in \cite{lu2023cnn}.
Recognizing the importance of local perception and long-range correlations, Xu et al. \cite{xu2021improved} added a Local Perception Block (LPB) to the Swin Transformer block in the Swin Transformer. This new backbone, called the Local Perception Swin Transformer (LPSW), improved the detection of small-size objects in aerial images significantly. DIAG-TR \cite{xue2022dual} introduced a Global-Local Feature Interweaving (GLFI) module in the encoder to adaptively and hierarchically embed local features into global representations. This technique counterbalances for the scale discrepancies of small objects. Furthermore, learnable anchor box coordinates were added to the content queries in the transformer decoder, providing an inductive bias.
In a recent study, Chen et al. \cite{chen2023hyneter} proposed the Hybrid network Transformer (Hyneter), which extends the range of local information by embedding convolutions into the transformer blocks. This improvement led to enhanced detection results on the MS COCO dataset. Similar hybrid approaches have been adopted in \cite{ding2023sw}. In another study \cite{yang2023unifying}, the authors proposed a new backbone called NeXtFormer, which combines CNN and transformer to boost the local details and features of small objects, while also providing a global receptive field.

Among various methods, O$^2$DETR \cite{ma2021oriented} substituted the attention mechanism in transformers with depthwise separable convolution. This change not only decreased memory usage and computational costs associated with multi-scale features but also potentially enhanced the detection accuracy in aerial photographs.

\begin{figure*}
\begin{center}
     \includegraphics[width=0.9\linewidth]{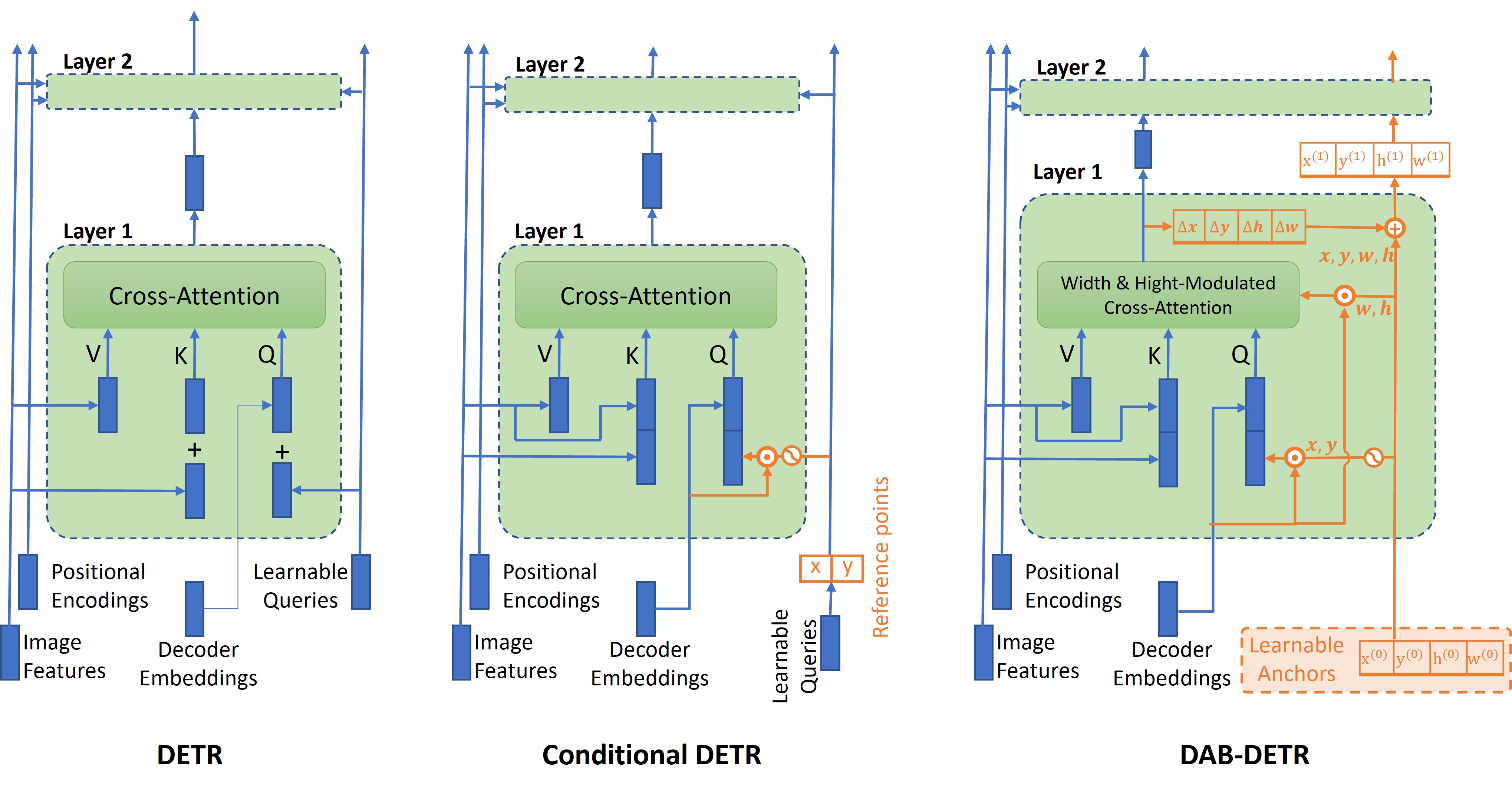}
\end{center}
   \caption{DAB-DETR improves Conditional DETR and utilizes dynamic  anchor boxes to sequentially provide better reference query points and anchor sizes (figure from \cite{liu2022dab}).}
\label{dab}
\end{figure*}

Questioning the object queries used in previous works, Wang et al. \cite{wang2022anchor} proposed Anchor DETR, which used anchor points for object queries. These anchor points enhance the interpretability of the target query locations. The use of multiple patterns for each anchor point, improves the detection of multiple objects in one region. In contrast, Conditional DETR \cite{meng2021conditional} emphasizes on the conditional spatial queries derived from the decoder content leading to spatial attention predictions. A subsequent version, Conditional DETR v2 \cite{chen2022conditional}, enhanced the architecture by reformulating the object query into the form of a box query. This modification involves embedding a reference point and transforming boxes with respect to the reference point. In subsequent works, DAB-DETR \cite{liu2022dab} further improved on the idea of query design by using dynamically adjustable anchor boxes. These anchor boxes serve as both reference query points and anchor dimensions (see Figure \ref{dab}).

In another work \cite{cao2022cf}, the authors observed that while the mean average precision (mAP) of small objects in DETR is not competitive with state-of-the-art (SOTA) techniques, its performance for small intersection-over-union (IoU) thresholds is surprisingly better than its competitors. This indicates that while DETR provides strong perception abilities, it requires fine-tuning to achieve better localization accuracy. As a solution, the Coarse-to-Fine Detection Transformer (CF-DETR) has been proposed to perform this refinement through Adaptive Scale Fusion (ASF) and Local Cross-Attention (LCA) modules in the decoder layer. In \cite{he2022destr} the authors contend that the suboptimal performance of transformer-based detectors can be attributed to factors such as using a singular cross-attention module for both categorization and regression, inadequate initialization for content queries, and the absence of leveraging prior knowledge in the self-attention module. To address these concerns, they proposed Detection Split Transformer (DESTR). This model splits cross-attention into two branches, one for classification and one for regression. Moreover, DESTR uses a mini-detector to ensure proper content query initialization in the decoder and enhances the self-attention module. Another research \cite{xu2022fea}, introduced FEA-Swin, which leverages advanced foreground enhancement attention in the Swin Transformer framework to integrate context information into the original backbone. This was motivated by the fact that Swin Transformer does not adequately handle dense object detection due to missing connections between adjacent objects. Therefore, foreground enhancement highlights the objects for further correlation analysis. TOLO \cite{xia2022transformers} is one of the recent works aiming to bring inductive bias (using CNN) to the transformer architecture through a simple neck module. This module combines features from different layers to incorporate high-resolution and high-semantic properties. Multiple light transformer heads were designed to detect objects at different scales. In a different approach, instead of modifying the modules in each architecture, CBNet, proposed by Liang et al. \cite{liang2022cbnet}, groups multiple identical backbones that are connected through composite connections.

In the Multi-Source Aggregation Transformer (MATR) \cite{sun2022multi}, the cross-attention module of the transformer is used to leverage other support images of the same object from different views. A similar approach is adopted in \cite{isaac2022multi}, where the Multi-View Vision Transformer (MVViT) framework combines information from multiple views, including the target view, to improve the detection performance when objects are not visible in a single view.

Other works prefer to adhere to the YOLO family architecture. For instance, SPH-Yolov5 \cite{gong2022swin} adds a new branch in the shallower layers of the Yolov5 network to fuse features for improved small object localization. It also incorporates for the first time the Swin Transformer prediction head in the Yolov5 pipeline.

In \cite{ouyang2022nms}, the authors argue that the Hungarian loss's direct one-to-one bounding box matching approach might not always be advantageous. They demonstrate that employing a one-to-many assignment strategy and utilizing the NMS (Non-Maximum Suppression) module leads to better detection results. Echoing this perspective, Group DETR \cite{chen2022group1} implements K groups of object queries with one-to-one label assignment, leading to K positive object queries for each ground-truth object to enhance performance.

A Dual-Key Transformer Network (DKTNet) is proposed in \cite{xu2023dktnet}, where two keys are used—one key along with the $\textbf{Q}$ stream and another key along with the $\textbf{V}$ stream. This enhances the coherence between $\textbf{Q}$ and $\textbf{V}$, leading to improved learning. Additionally, channel attention is computed instead of spatial attention, and 1D convolution is used to accelerate the process.

\subsection{Auxiliary Techniques}
Experimental results have demonstrated that auxiliary techniques or tasks, when combined with the main task, can enhance performance. In the context of transformers, several techniques have been adopted, including: \textbf{(i)} Auxiliary Decoding/Encoding Loss: This refers to the approach where feed-forward networks designed for bounding box regression and object classification are connected to separate decoding layers. Hence individual losses at different scales are combined to train the models leading to better detection results. This technique or its variants have been used in ViDT \cite{song2022vidt}, MDef-DETR \cite{maaz2021multi}, CBNet \cite{liang2022cbnet}, SRDD \cite{zhu2022srdd}.  \textbf{(ii)} Iterative Box Refinement: In this method, the bounding boxes within each decoding layer are refined based on the predictions from the previous layers. This feedback mechanism progressively improves detection accuracy. This technique has been used in  ViDT \cite{song2022vidt}. \textbf{(iii)} Top-Down Supervision: This approach leverages human understandable semantics to aid in the intricate task of detecting small or class-agnostic objects, e.g., aligned image-text pairs in MDef-DETR \cite{maaz2021multi}, or text-guided object detector in TGOD \cite{shen2023text}. \textbf{(iv)} Pre-training: This involves training on large-scale datasets followed by specific fine-tuning for the detection task. This technique has been used in CBNet V2-TTA \cite{cai2022bigdetection}, FP-DETR \cite{wang2022fp}, T-TRD \cite{li2022transformer}, SPH-Yolov5 \cite{gong2022swin}, MATR \cite{sun2022multi}, and extensively in Group DETR v2 \cite{chen2022group}. \textbf{(v)} Data Augmentation: This technique enriches the detection dataset by applying various augmentation techniques, such as rotation, flipping, zooming in and out, cropping, translation, adding noise, etc. Data augmentation is a commonly used approach to address various imbalance problems \cite{oksuz2020imbalance}, e.g., imbalance in object size, within deep learning datasets. Data augmentation can be seen as an indirect approach to minimize the gap between train and test sets \cite{rashidi2022ruda}. Several methods used augmentation in their detection task including T-TRD \cite{li2022transformer}, SPH-Yolov5 \cite{gong2022swin}, MATR \cite{sun2022multi}, NLFFTNet \cite{zeng2022nlfftnet}, DeoT \cite{ding2023deot}, HTDet \cite{chen2023htdet}, and Sw-YoloX \cite{ding2023sw}. (\textbf{vi}) One-to-Many Label Assignment: The one-to-one matching in DETR can result in poor discriminative features within the encoder. Hence, one-to-many assignments in other methods, e.g., Faster-RCNN, RetinaNet, and FCOS have been used as auxiliary heads in some studies such as CO-DETR \cite{zong2022detrs}. (\textbf{vii}) Denoising Training: This technique aims to boost the convergence speed of the decoder in DETR, which often faces an unstable convergence due to bipartite matching. In denoising training, the decoder is fed with noisy ground-truth 
labels and boxes into the decoder. The model is then trained to reconstruct the original ground truth (guided by an auxiliary loss). Implementations like DINO \cite{zhang2022dino} and DN-DETR \cite{li2022dn} have demonstrated the effectiveness of this technique in enhancing the decoder's stability.

\subsection{Improved Feature Representation}
Although current object detectors excel in a wide range of applications for regular-size or large objects, certain use-cases necessitate specialized feature representations for improved SOD. For instance, when it comes to detecting oriented objects in aerial imagery, any object rotation can drastically alter the feature representation due to increased background noise or clutter in the scene (region proposal). To address this, Dai~\textit{et al.} \cite{dai2022ao2} ] proposed AO2-DETR, a method designed to be robust to arbitrary object rotations. This is achieved through three key components: \textbf{(i)} the generation of oriented proposals,  \textbf{(ii)} a refinement module of the oriented proposal which extracts rotational-invariant features, and \textbf{(iii)} a rotation-aware set matching loss. These modules help to negate the effects of any rotations of the objects.  In a related approach, DETR++\cite{zhang2022detr++}, uses multiple Bi-Directional Feature Pyramid layers (BiFPN) that are applied in a bottom-up fashion to feature maps from C3, C4, and C5. Then, only one scale which is representative of features at all scales is selected to be fed into DETR framework for detection. For some specific applications, such as plant safety monitoring, where objects of interest are usually related to human workers, leveraging this contextual information can greatly improve feature representation. PointDet++ \cite{tang2022pointdet++} capitalizes on this by incorporating human pose estimation techniques, integrating local and global features to enhance SOD performance. Another crucial element that impacts feature quality is the backbone network and its ability to extract both semantic and high-resolution features.  GhostNet introduced in \cite{li2022ghostformer}, offers a streamlined and more efficient network that delivers high-quality, multi-scale features to the transformer. Their Ghost module in this network partially generates the output feature map, with the remainder being recovered using simple linear operations. This is a key step to alleviate the complexity of the backbone networks. 
In the context of medical image analysis, MS Transformer \cite{shou2022object} used a self-supervised learning approach  to perform a random mask on the input image, which aids in reconstructing richer features, that are less sensitive to the noise. In conjunction with a hierarchical transformer, this approach outperforms DETR frameworks with various backbones. The Small Object Favoring DETR (SOF-DETR) \cite{dubey2022improving}, specifically favors the detection of small objects by merging convolutional features from layers 3 and 4 in a  normalized inductive bias module prior to input into the DETR-Transformer. NLFFTNet \cite{zeng2022nlfftnet} addresses the limitation of only considering local interactions in current fusion techniques by introducing a nonlocal feature-fused transformer convolutional network, capturing long-distance semantic relationships between different feature layers. DeoT \cite{ding2023deot} merges an encoder-only transformer with a novel feature pyramid fusion module. This fusion is enhanced by the use of channel and spatial attention in the Channel Refinement Module (CRM) and Spatial Refinement Module (SRM), enabling the extraction of richer features. The authors in HTDet \cite{chen2023htdet} proposed a fine-grained FPN to cumulatively fuse low-level and high-level features for better object detection. Meanwhile, in MDCT \cite{chen2023mdct} the author proposed a Multi-kernel Dilated Convolution (MDC) module to improve the performance of small
object-related feature extraction using both the ontology and adjacent spatial features of small objects. The proposed module leverages depth-wise separable convolution to reduce the computational cost. Lastly, in \cite{ye2023real}, a feature fusion module paired with a lightweight backbone is engineered to enhance the visual features of small objects by broadening the receptive field. The hybrid attention module in RTD-Net  \cite{ye2023real} empowers the system to detect objects that are partially occluded, by incorporating contextual information surrounding small objects.
\begin{figure*}
\begin{center}
   \includegraphics[width=1\linewidth]{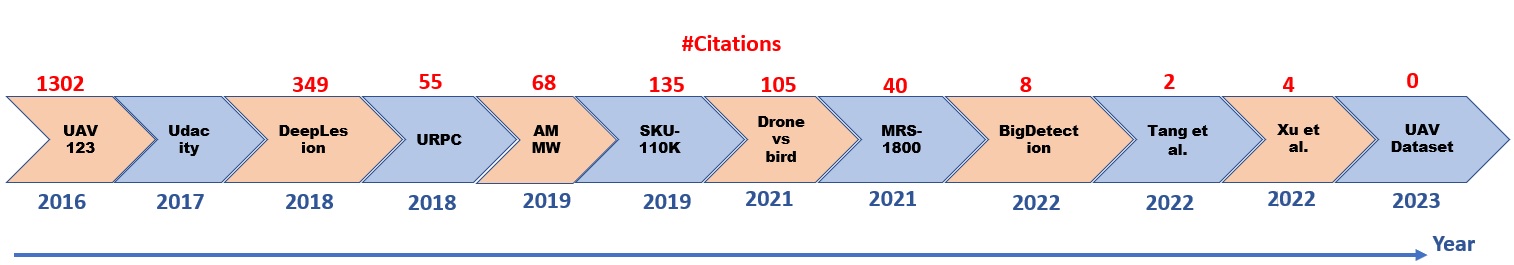}
\end{center}
   \caption{Chronology of SOD datasets with number of citations (based on Google Scholar).}
\label{datasets}
\end{figure*}
\subsection{Spatio-Temporal Information}
In this section, our focus is exclusively on video-based object detectors that aim to identify small objects. While many of these studies have been tested on the ImageNet VID dataset \footnote{\url{https://paperswithcode.com/sota/video-object-detection-on-imagenet-vid} } \cite{russakovsky2015imagenet}, this dataset was not originally intended for small object detection.  Nonetheless, a few of the works also reported their results for small objects of ImageNet VID dataset.   The topic of tracking and detecting small objects in videos has also been explored using transformer architectures. Although techniques for image-based SOD can be applied to video, they generally do not utilize the valuable temporal information, which can be particularly beneficial for identifying small objects in cluttered or occluded frames. The application of transformers to generic object detection/tracking started with  TrackFormer \cite{meinhardt2022trackformer} and TransT \cite{chen2021transformer}. These models used frame-to-frame (setting the previous frame as the reference) set prediction and template-to-frame (setting a template frame as the reference) detection. Liu \textit{et al.} in \cite{liu2021aerial} were among the first to use transformers specifically for video-based small object detection and tracking. Their core concept is to update template frames to capture any small changes induced by the presence of small objects and to provide a global attention-driven relationship between the template frame and the search frame.\\
Transformer-based object detection gained formal recognition with the introduction of TransVOD, an end-to-end object detector, as presented in  \cite{he2021end} and \cite{zhou2022transvod}. This model applies both spatial and temporal transformers to a series of video frames, thereby identifying and linking objects across these frames. TransVOD has spawned several variants, each with unique features, including capabilities for real-time detection. PTSEFormer \cite{wang2022ptseformer} adopts a progressive strategy, focusing on both temporal information and the objects’ spatial transitions between frames. It employs multi-scale feature extraction to achieve this. Unlike other models, PTSEFormer directly regresses object queries from adjacent frames rather than the entire dataset, offering a more localized approach.\\
Sparse VOD \cite{hashmi2022spatio} proposed an end-to-end trainable video object detector that incorporates temporal information to propose region proposals. In contrast, DAFA \cite{roh2022dafa} highlights the significance of global features within a video as opposed to local temporal features. DEFA showed the inefficiency of the First In First Out (FIFO) memory structure and proposed a diversity-aware memory, which uses object-level memory instead of frame-level memory for the attention module.   VSTAM \cite{fujitake2022video} improves feature quality on an element-by-element basis and then performs sparse aggregation before these enhanced features are used for object candidate region detection. The model also incorporates external memory to take advantage of long-term contextual information.\\
In the FAQ work \cite{cui2023faq}, a novel video object detector is proposed that uses query feature aggregation in the decoder module. This is different than the methods that focus on either feature aggregation in the encoder or the methods that perform post-processing for various frames. The research indicates that this technique improves the detection performance outperforming SOTA methods.

\section{Results and Benchmarks}\label{Results}
In this section, we quantitatively and qualitatively evaluate previous works of small object detection, identifying the most effective technique for a specific application. Prior to this comparison, we introduce a range of new datasets dedicated to small object detection, including both videos and images for diverse applications.
\begin{table*}
	\caption{Commonly used datasets for SOD. NF: Not fixed.}
	\label{tab2}
	\centering
	\scalebox{0.7}{\begin{tabular}{|c||c|c|c|c|c|c|c|c|p{2cm}|}
		\hline
		\textbf{Dataset} & \textbf{Application} & \textbf{Video}&\textbf{Image}  & \textbf{Shooting Angle (Type)} & \textbf{Resolution (pixels)}& \#\textbf{Object Classes} & \#\textbf{Instances} & \#\textbf{Images/Videos}&\textbf{Public?}  \\
		\hline
		\hline	
			\textbf{UAV123} \cite{mueller2016benchmark} & UAV Tracking & \checkmark  &  & Aerial Perspective(RGB) & -- &-- &--   &123 ($>$110K frames) & Yes: \href{https://cemse.kaust.edu.sa/ivul/uav123}{\textcolor{blue}{Click Here}}  \\
			\hline

   			\textbf{MRS-1800} \cite{xu2021improved} & Remote Sensing &   & \checkmark & Satellite based(RGB)& NF & 3&  16,318 &1800 & -- \\
			\hline

      \textbf{SKU-110K}\cite{goldman2019precise}& Commodity Detection & & \checkmark& Normal & NF & 110,712& 147.4 per image& 11,762& Yes: \href{https://github.com/eg4000/SKU110K_CVPR19}{\textcolor{blue}{Click Here}}  \\
      \hline
\textbf{BigDetection}\cite{cai2022bigdetection}& Generic& & \checkmark&Normal& NF&600 &36M &\makecell{3.4M Training\\141K Test} &Yes: \href{https://github.com/amazon-science/bigdetection}{\textcolor{blue}{Click Here}} \\
\hline

\textbf{Tang et al.} \cite{tang2022pointdet++}& Cemical Plant Monitoring& & \checkmark & Normal& --& 19& --& 2400& --\\
\hline

\textbf{Xu et al.} \cite{xu2022fea}& UAV-based Detection & & \checkmark & Aerial (RGB)&1920$\times$1080 &2&12.5K & 2K&Yes: \href{https://pan.baidu.com/share/init?surl=4UcfTtZnvvVyCV2tAzHFKw}{\textcolor{blue}{Click Here}}\\
\hline
\textbf{DeepLesion} \cite{yan2018deeplesion}& Lesion detection & & \checkmark & (CT)&--&8&32.7K &32.1K&Yes: \href{https://nihcc.app.box.com/v/DeepLesion}{\textcolor{blue}{Click Here}}\\
\hline
\textbf{Udacity Self Driving Car} \cite{udacity}& Self-Driving& \checkmark& &Normal& 1920$\times$1200& 3& 65K& 9,423& Yes: \href{https://github.com/udacity/self-driving-car/tree/master/annotations}{\textcolor{blue}{Click Here}}\\
      \hline

 \textbf{AMMW Dataset} \cite{liu2019concealed}& Security Inspection& & \checkmark & Normal (AMMW)& 160$\times$400&$>$30 &-- & $>$58K&  --  \\
\hline
 \textbf{URPC 2018 Dataset} \footnote{http://en.cnurpc.org/} & Underwater Detection& & \checkmark& Normal& --& 4&-- & \makecell{2,901 Training\\800  Test}&--\\

 \hline
\textbf{UAV dataset} \cite{ye2023real}& UAV-based detection& & \checkmark& Aerial Perspective (RGB)&-- &7 & 320,624& 9,630& --\\
 \hline
 \textbf{Drone-vs-bird} \cite{coluccia2021drone}& Drone Detection& \checkmark& &Normal&NF &2 &-- & 77 Training 1,384 Frames&Yes: \href{https://github.com/wosdetc/challenge}{\textcolor{blue}{Click Here}} \\
 \hline
	\end{tabular}}
\end{table*}
\subsection{Datasets}
In this subsection, in addition to the widely used MS COCO dataset, we compile and present 12 new SOD datasets. These new datasets are primarily tailored for specific applications excluding the generic and maritime environments (which have been covered in our previous survey \cite{rekavandi2022guide}). Figure \ref{datasets} displays the chronological order of these datasets along with their citation count as of June 15 2023, according to Google Scholar. \\
\textbf{UAV123} \cite{mueller2016benchmark}: This dataset contains 123 videos acquired with UAVs and it is one of the largest object-tracking datasets with more than 110K frames. \\
\textbf{MRS-1800} \cite{xu2021improved}: ]: This dataset consists of a combination of images from three other remote sensing datasets: DIOR \cite{li2020object}, NWPU VHR-10 \cite{cheng2016learning}, and HRRSD \cite{zhang2019hierarchical}. MRD-1800 was created for the dual purpose of detection and instance segmentation, with 1800 manually annotated images which include 3 types of objects: airplanes, ships, and storage tanks. \\
\textbf{SKU-110K \cite{goldman2019precise}}: This dataset serves as a rigorous testbed for commodity detection, featuring images captured from various supermarkets around the world. The dataset includes a range of scales, camera angles, lighting conditions, etc.\\
\textbf{BigDetection}\cite{cai2022bigdetection}: This is a large-scale dataset that is crafted by integrating existing datasets and meticulously eliminating duplicate boxes while labeling overlooked objects. It has a balanced number of objects across all sizes making it a pivotal resource for advancing the field object detection. Using this dataset for pretraining and subsequently fine-tuning on MS COCO significantly enhances performance outcomes. \\
\textbf{Tang et al.} \cite{tang2022pointdet++}: Originating from video footage of field activities within a chemical plant, this dataset covers various types of work such as hot work, aerial work, confined space operations, etc. It includes category labels like people, helmets, fire extinguishers, gloves, work clothes and other relevant objects.\\ 
\begin{table*}
\caption{Detection performance (\%) for small-scale objects on MS COCO image dataset \cite{lin2014microsoft}. The top section shows results for CNN-based techniques, the middle section shows results for mixed architectures, and the bottom section presents from transformer-only networks. DC5: Dilated C5 stage, MS: Multi-scale network, IBR: Iterative bounding box refinement, TS: Two-stage detection, DCN: Deformable convnets, TTA: Test time augmentation, BD: Pre-trained on BigDetection dataset, IN: Pre-trained on ImageNet, OB: Pre-trained on Object-365 \cite{gao2019solution}. While ${}^*$ shows the results for COCO test-dev, the other values are reported for COCO val set.}
\centering
\label{tab5}
\scalebox{0.97}{\begin{tabular}{lcccccc}
\toprule
Model& Backbone& GFLOPS$\downarrow$/FPS $\uparrow$ & $\#$params$\downarrow$ &$\text{mAP}^{@[0.5,0.95]}$  $\uparrow$& Epochs$\downarrow$&URL  \\ \hline
Faster RCNN-DC5~(NeurIPS2015)\cite{ren2015faster}& ResNet50&320/16 & 166M&  21.4& 37&\href{https://github.com/trzy/FasterRCNN}{\textcolor{blue}{Link}}\\ 
  
Faster RCNN-FPN~(NeurIPS2015)\cite{ren2015faster}& ResNet50&180/26 & 42M& 24.2&37&\href{https://github.com/trzy/FasterRCNN}{\textcolor{blue}{Link}}\\ 
 
Faster RCNN-FPN~(NeurIPS2015)\cite{ren2015faster}& ResNet101&246/20& 60M &  25.2&--&\href{https://github.com/trzy/FasterRCNN}{\textcolor{blue}{Link}}\\ 

RepPoints v2-DCN-MS~(NeurIPS2020)\cite{chen2020reppoints}& ResNeXt101&--/--&--&34.5$^*$&24&\href{https://github.com/Scalsol/RepPointsV2}{\textcolor{blue}{Link}}\\

FCOS~(ICCV2019)\cite{tian2019fcos}& ResNet50& 177/17&--&26.2&36&\href{https://github.com/tianzhi0549/FCOS}{\textcolor{blue}{Link}}\\

CBNet V2-DCN(ATSS\cite{zhang2020bridging})~(TIP2022)\cite{liang2022cbnet}&Res2Net101 &--/--&107M&35.7$^*$&20&\href{https://github.com/VDIGPKU/CBNetV2}{\textcolor{blue}{Link}}\\

CBNet V2-DCN(Cascade RCNN)~(TIP2022)\cite{liang2022cbnet}& Res2Net101&--/--&146M&37.4$^{*}$&32&\href{https://github.com/VDIGPKU/CBNetV2}{\textcolor{blue}{Link}}\\

\hline

DETR~(ECCV2020)\cite{carion2020end}& ResNet50&86/\textbf{28} & 41M & 20.5&500&\href{https://github.com/facebookresearch/detr}{\textcolor{blue}{Link}}\\

DETR-DC5~(ECCV2020)\cite{carion2020end}& ResNet50&187/12& 41M&22.5&500&\href{https://github.com/facebookresearch/detr}{\textcolor{blue}{Link}}\\

DETR~(ECCV2020)\cite{carion2020end}& ResNet101&\textbf{52}/20& 60M&21.9&--&\href{https://github.com/facebookresearch/detr}{\textcolor{blue}{Link}}\\

DETR-DC5~(ECCV2020)\cite{carion2020end}&ResNet101&253/10 & 60M& 23.7&--&\href{https://github.com/facebookresearch/detr}{\textcolor{blue}{Link}}\\

ViT-FRCNN~(arXiv2020)\cite{beal2020toward}&--&--/-- & --& 17.8&--&--\\

RelationNet++~(NeurIPS2020)\cite{chi2020relationnet++}&ResNeXt101&--/-- & --& 32.8$^{*}$&--&\href{https://github.com/microsoft/RelationNet2}{\textcolor{blue}{Link}}\\

RelationNet++-MS~(NeurIPS2020)\cite{chi2020relationnet++}&ResNeXt101&--/-- & --& 35.8$^*$&--&\href{https://github.com/microsoft/RelationNet2}{\textcolor{blue}{Link}}\\

Deformable DETR~(ICLR2021)\cite{zhu2021deformable}& ResNet50& 173/19& 40M&26.4&50& \href{https://github.com/fundamentalvision/Deformable-DETR}{\textcolor{blue}{Link}}\\

Deformable DETR-IBR~(ICLR2021)\cite{zhu2021deformable}& ResNet50& 173/19& 40M& 26.8& 50&\href{https://github.com/fundamentalvision/Deformable-DETR}{\textcolor{blue}{Link}}\\

Deformable DETR-TS~(ICLR2021)\cite{zhu2021deformable}& ResNet50&173/19 &40M &28.8& 50&\href{https://github.com/fundamentalvision/Deformable-DETR}{\textcolor{blue}{Link}}\\

Deformable DETR-TS-IBR-DCN~(ICLR2021)\cite{zhu2021deformable}&ResNeXt101&--/-- &--  &34.4$^{*}$& --&\href{https://github.com/fundamentalvision/Deformable-DETR}{\textcolor{blue}{Link}}\\

Dynamic DETR~(ICCV2021)\cite{dai2021dynamic} &ResNet50&--/-- &-- &28.6$^*$& --&--\\

Dynamic DETR-DCN~(ICCV2021)\cite{dai2021dynamic} &ResNeXt101&--/-- &-- &30.3$^*$& --&--\\

TSP-FCOS~(ICCV2021)\cite{sun2021rethinking}&ResNet101&255/12&--&27.7&36&\href{https://github.com/Edward-Sun/TSP-Detection}{\textcolor{blue}{Link}}\\

TSP-RCNN~(ICCV2021)\cite{sun2021rethinking}&ResNet101&254/9&--&29.9&96&\href{https://github.com/Edward-Sun/TSP-Detection}{\textcolor{blue}{Link}}\\

Mask R-CNN~(ICCV2021)\cite{peng2021conformer}&Conformer-S/16& 457.7/--& 56.9M&28.7&\textbf{12}&\href{https://github.com/pengzhiliang/Conformer}{\textcolor{blue}{Link}}\\

Conditional DETR-DC5~(ICCV2021)\cite{meng2021conditional}&ResNet101 &262/--& 63M& 27.2&108 & \href{https://github.com/Atten4Vis/ConditionalDETR}{\textcolor{blue}{Link}}\\

SOF-DETR~(2022JVCIR) \cite{dubey2022improving}& ResNet50 & --/--& --& 21.7& --& \href{https://github.com/shikha-gist/SOF-DETR/}{\textcolor{blue}{Link}}\\

DETR++~(arXiv2022)\cite{zhang2022detr++}& ResNet50 & --/--&-- & 22.1&-- &--\\

TOLO-MS~(NCA2022) \cite{xia2022transformers}& --& --/57&--&24.1&--&--\\

Anchor DETR-DC5~(AAAI2022) \cite{wang2022anchor} &ResNet101&--/--&--&25.8&50& \href{https://github.com/megvii-research/AnchorDETR}{\textcolor{blue}{Link}}\\

DESTR-DC5~(CVPR2022)\cite{he2022destr}& ResNet101& 299/--&88M&28.2&50&--\\

Conditional DETR v2-DC5~(arXiv2022)\cite{chen2022conditional}&ResNet101 &228/--& 65M& 26.3&50 & --\\

Conditional DETR v2~(arXiv2022)\cite{chen2022conditional}&Hourglass48 &521/--& 90M& 32.1&50 &-- \\

FP-DETR-IN~(ICLR2022) \cite{wang2022fp} & --& --/--&\textbf{36M} &26.5 & 50&\href{https://github.com/encounter1997/FP-DETR}{\textcolor{blue}{Link}}\\

DAB-DETR-DC5~(arXiv2022)\cite{liu2022dab}& ResNet101& 296/--& 63M& 28.1& 50& \href{https://github.com/IDEA-Research/DAB-DETR}{\textcolor{blue}{Link}} \\

Ghostformer-MS~(Sensors2022)\cite{li2022ghostformer}& GhostNet& --/--& --& 29.2& 100&-- \\

CF-DETR-DCN-TTA~(AAAI2022)\cite{cao2022cf}&ResNeXt101 &--/-- & --&35.1$^*$&--&--\\

CBNet V2-TTA~(CVPR2022)\cite{cai2022bigdetection}&Swin Transformer-base &--/--&--&41.7&--&\href{https://github.com/amazon-science/bigdetection}{\textcolor{blue}{Link}}\\

CBNet V2-TTA-BD~(CVPR2022)\cite{cai2022bigdetection}&Swin Transformer-base &--/--&--&42.2&--&\href{https://github.com/amazon-science/bigdetection}{\textcolor{blue}{Link}}\\

DETA~(arXiv2022)\cite{ouyang2022nms}&ResNet50& --/13&48M& 34.3&24&\href{https://github.com/jozhang97/DETA}{\textcolor{blue}{Link}}\\

DINO~(arXiv2022)\cite{zhang2022dino}& ResNet50&860/10& 47M& 32.3& \textbf{12}&  \href{https://github.com/IDEA-Research/DINO}{\textcolor{blue}{Link}}\\

CO-DINO Deformable DETR-MS-IN~(arXiv2022)\cite{zong2022detrs}& Swin Transformer-large&--/--&--& 43.7&36&\href{https://github.com/Sense-X/Co-DETR}{\textcolor{blue}{Link}}\\

HYNETER~(ICASSP2023)\cite{chen2023hyneter}& Hyneter-Max& --/--&247M&29.8$^{*}$&--&-- \\

DeoT~(JRTIP2023) \cite{ding2023deot}& ResNet101 & 217/14 & 58M& 31.4& 34&-- \\

ConformerDet-MS~(TPAMI2023) \cite{peng2023conformer} & Conformer-B & --/-- & 147M&35.3 & 36 & \href{https://github.com/pengzhiliang/Conformer}{\textcolor{blue}{Link}}\\
\hline

YOLOS~(NeurIPS2021)\cite{fang2021you}&DeiT-base&--/3.9&100M&19.5&150&\href{https://github.com/hustvl/YOLOS}{\textcolor{blue}{Link}}\\

DETR(ViT)~(arXiv2021)\cite{song2022vidt} &Swin Transformer-base&--/9.7&100M&18.3&50&\href{https://github.com/naver-ai/vidt}{\textcolor{blue}{Link}}\\

Deformable DETR(ViT)~(arXiv2021)\cite{song2022vidt} &Swin Transformer-base&--/4.8&100M&34.5&50&\href{https://github.com/naver-ai/vidt}{\textcolor{blue}{Link}}\\

ViDT~(arXiv2022)\cite{song2022vidt}&Swin Transformer-base&--/9&100M&30.6&50&\href{https://github.com/naver-ai/vidt/tree/main}{\textcolor{blue}{Link}}\\

DFFT~(ECCV2022) \cite{chen2022efficient}& DOT-medium& 67/--& --& 25.5& 36& \href{https://github.com/PeixianChen/DFFT}{\textcolor{blue}{Link}}\\

CenterNet++-MS~(arXiv2022) \cite{duan2022centernet++}& Swin Transformer-large &--/--&--&38.7$^{*}$&--& \href{https://github.com/Duankaiwen/PyCenterNet}{\textcolor{blue}{Link}}\\

DETA-OB~(arXiv2022)\cite{ouyang2022nms}&Swin Transformer-large& --/4.2&--& 46.1$^{*}$&24&\href{https://github.com/jozhang97/DETA}{\textcolor{blue}{Link}}\\

Group DETR v2-MS-IN-OB~(arXiv2022) \cite{chen2022group}& ViT-Huge& --/--&629M& \textbf{48.4}$^{*}$&--&-- \\
\hline
Best Results&NA&DETR&FP-DETR&Group DETR v2&DINO&NA\\
\bottomrule
\end{tabular}}
\end{table*}
\textbf{Xu et al.} \cite{xu2022fea}: This publicly available dataset focuses on UAV (Unmanned Aerial Vehicle)-captured images and contains 2K images aimed at detecting both pedestrians and vehicles. The images were collected using a DJI drone and feature diverse conditions such as varying light levels and densely parked vehicles.\\
\textbf{DeepLesion} \cite{yan2018deeplesion}: Comprising CT scans from 4,427 patients, this dataset ranks among the largest of its kind. It includes a variety of lesion types, such as pulmonary nodules, bone abnormalities, kidney lesions, and enlarged lymph nodes. The objects of interest in these images are typically small and accompanied by noise, making their identification challenging.\\
\textbf{Udacity Self Driving Car} \cite{udacity}: Designed solely for educational use, this dataset features driving scenarios in Mountain View and nearby cities captured at a 2Hz image acquisition rate. The category labels within this dataset include cars, trucks, and pedestrians.\\
\textbf{AMMW Dataset} \cite{liu2019concealed}: Created for security applications, this active millimetre-wave image dataset includes more than 30 different types of objects. These include two kinds of lighters (made of plastic and metal), a simulated firearm, a knife, a blade, a bullet shell, a phone, a soup, a key, a magnet, a liquid bottle, an absorbent material, a match, and so on.\\
\textbf{URPC 2018 Dataset}: This underwater image dataset includes four types of objects: holothurian, echinus, scallop and starfish \cite{lin2020roimix}. \\
\textbf{UAV dataset} \cite{ye2023real}: This image dataset includes more than 9K images captured via UAVs in different weather and lighting conditions and various complex backgrounds. The objects in this dataset are sedans, people, motors, bicycles, trucks, buses, and tricycles. \\ 
\textbf{Drone-vs-bird} \cite{coluccia2021drone}: This video dataset aims to address the security concerns of drones flying over sensitive areas. It offers labeled video sequences to differentiate between birds and drones under various illumination, lighting, weather, and background conditions. \\
A summary of these datasets, including their applications, type, resolutions, number of classes/instances/images/frame, and a link to their webpage, is provided in Table \ref{tab2}.

\subsection{Benchmarks in Vision Applications}
In this subsection, we introduce various vision-based applications where the detection performance of small objects is vital. For each application, we select one of the most popular datasets and report its performance metrics, along with details of the experimental setup.

\subsubsection{Generic Applications}
For generic applications, we evaluate the performance of all small object detectors on the challenging MS COCO benchmark \cite{lin2014microsoft}. The choice of this dataset is based on its wide acceptance in the object detection field and the accessibility of performance results. The MS COCO dataset consists of approximately 160K images across 80 categories. While the authors are advised to train their algorithms using the COCO 2017 training and validation sets, they are not restricted to these subsets.

\begin{figure*}
	 \centering
	\scalebox{0.97}{\begin{tabular}{p{0.3cm} p{4.5cm} p{4cm} p{4cm} p{4cm}}
		   \tiny \textbf{Input} &   \includegraphics[width=1.07\linewidth]{Figures/000000037689.JPG} &\includegraphics[width=1.07\linewidth]{Figures/000000166287.JPG}
		 & \includegraphics[width=1.07\linewidth]{Figures/000000256775.jpg}& \includegraphics[width=1.07\linewidth]{Figures/000000350023.jpg}\\
		\hline
  \tiny \textbf{CBNet-V2}& \includegraphics[width=1.07\linewidth]{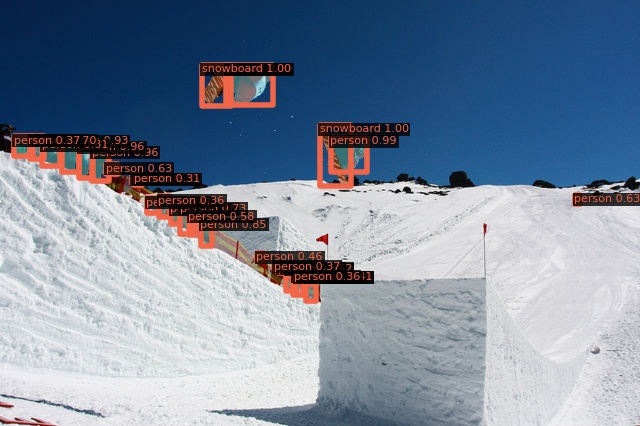} &\includegraphics[width=1.07\linewidth]{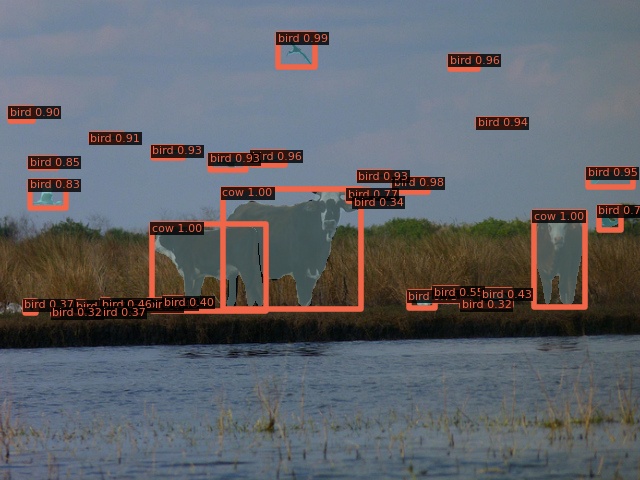}
		 & \includegraphics[width=1.07\linewidth]{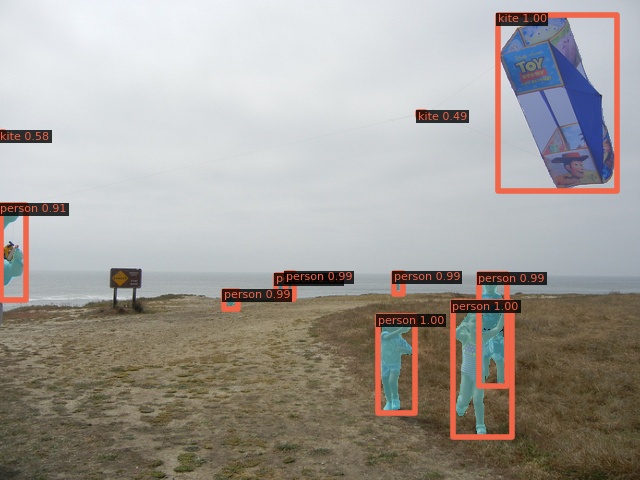}& \includegraphics[width=1.07\linewidth]{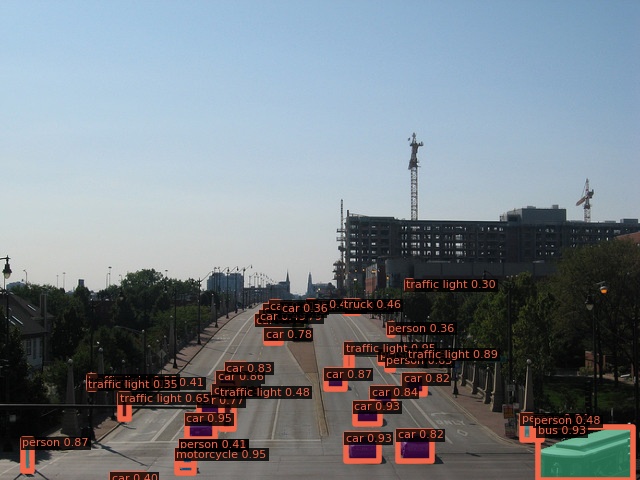}\\
   \hline
     \tiny\textbf{ DETA-OB}& \includegraphics[width=1.07\linewidth]{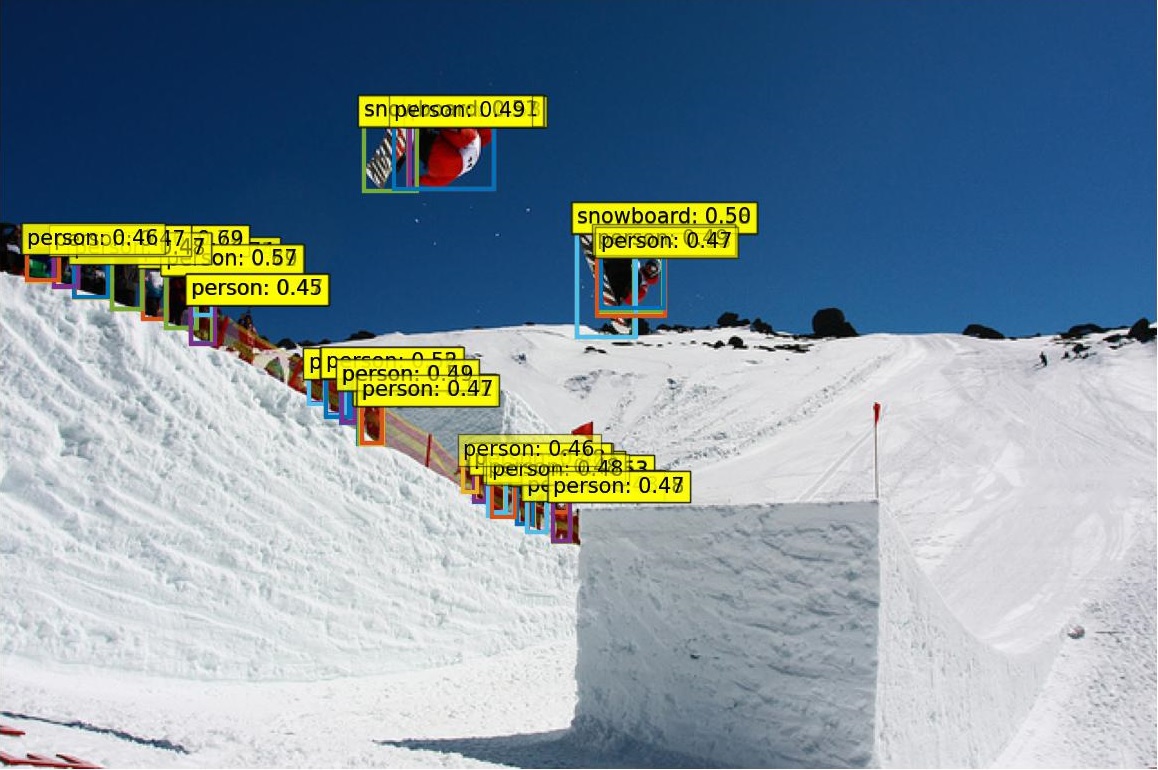} &\includegraphics[width=1.07\linewidth]{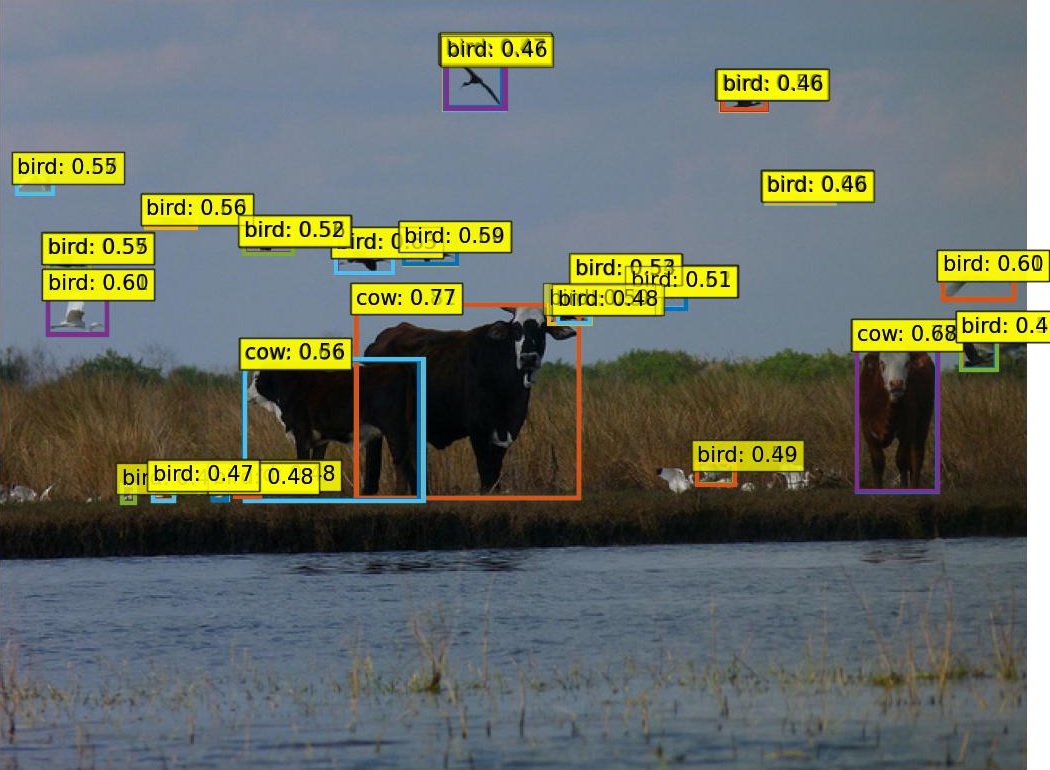}
		 & \includegraphics[width=1.07\linewidth]{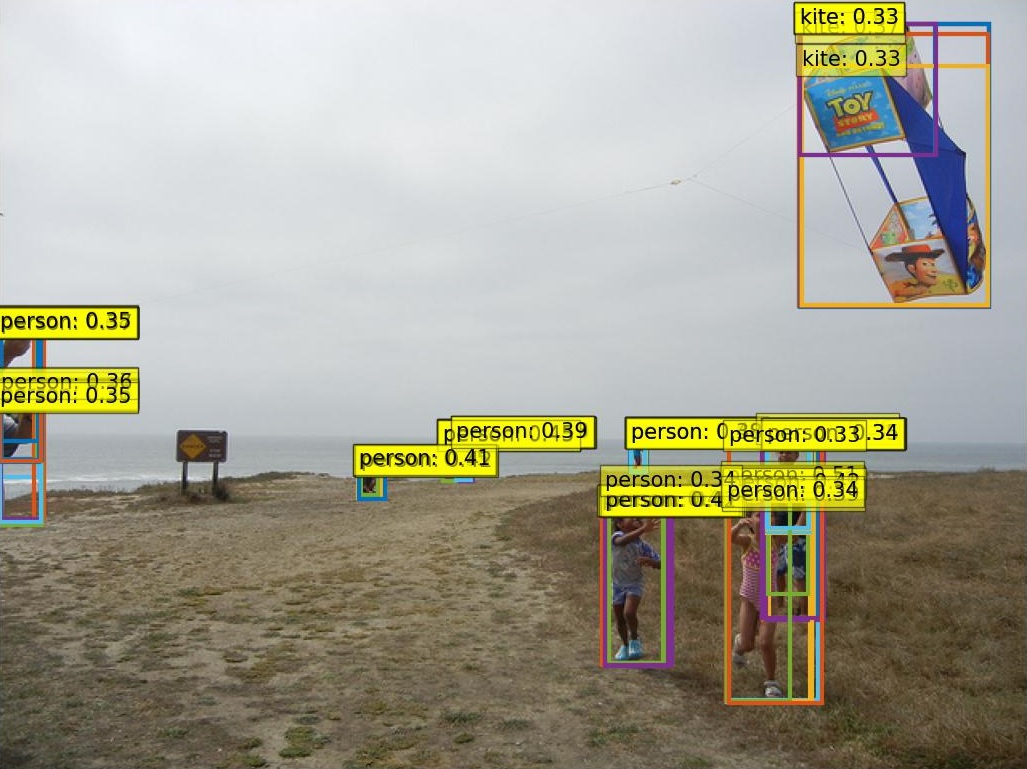}& \includegraphics[width=1.07\linewidth]{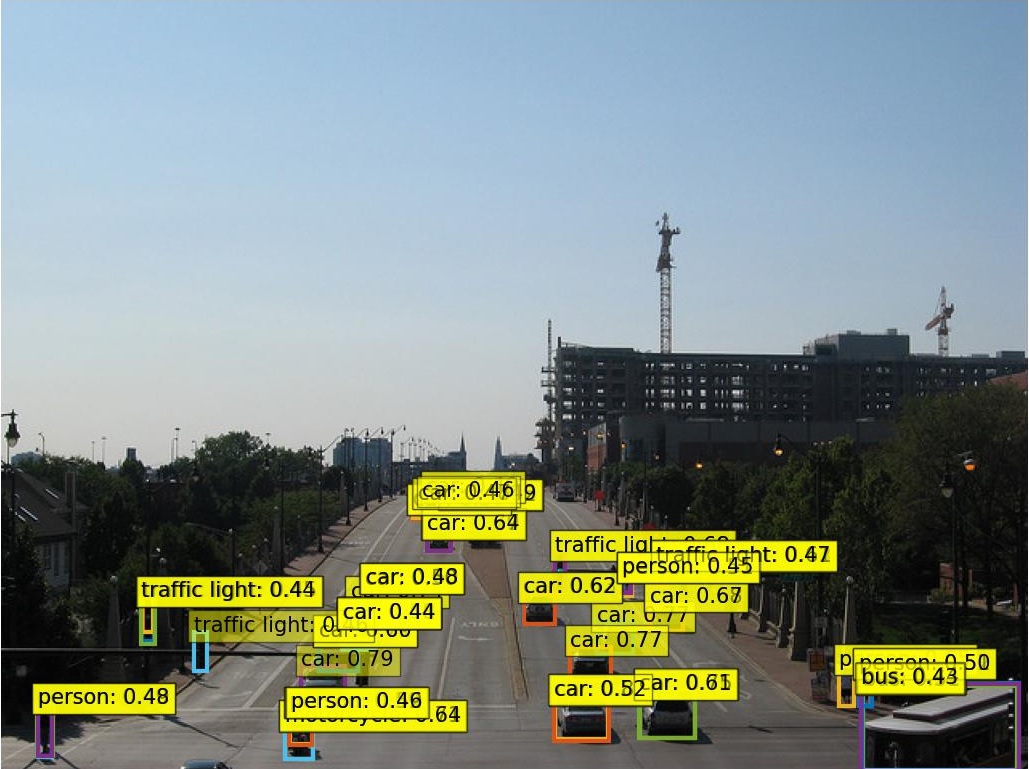}\\
   \hline
    \tiny \textbf{DINO}& \includegraphics[width=1.07\linewidth]{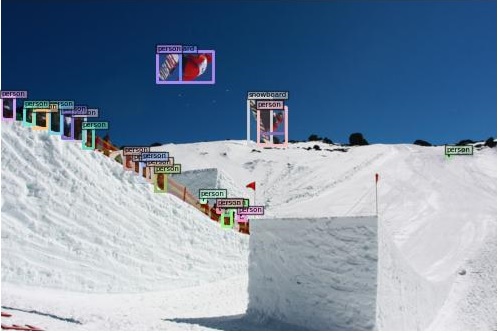} &\includegraphics[width=1.07\linewidth]{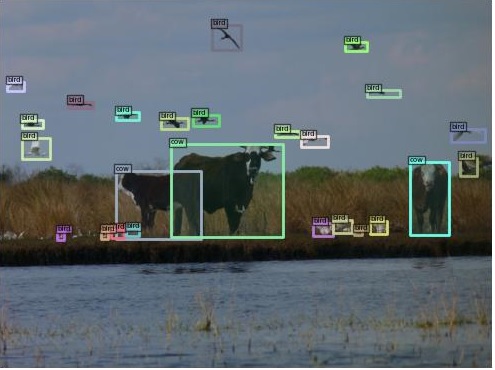}
		 & \includegraphics[width=1.07\linewidth]{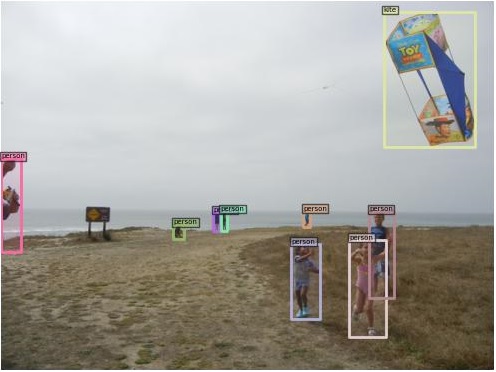}& \includegraphics[width=1.07\linewidth]{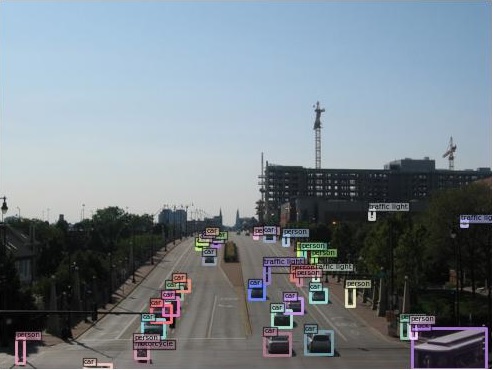}\\
   \hline
    \tiny \textbf{ViDT}& \includegraphics[width=1.07\linewidth]{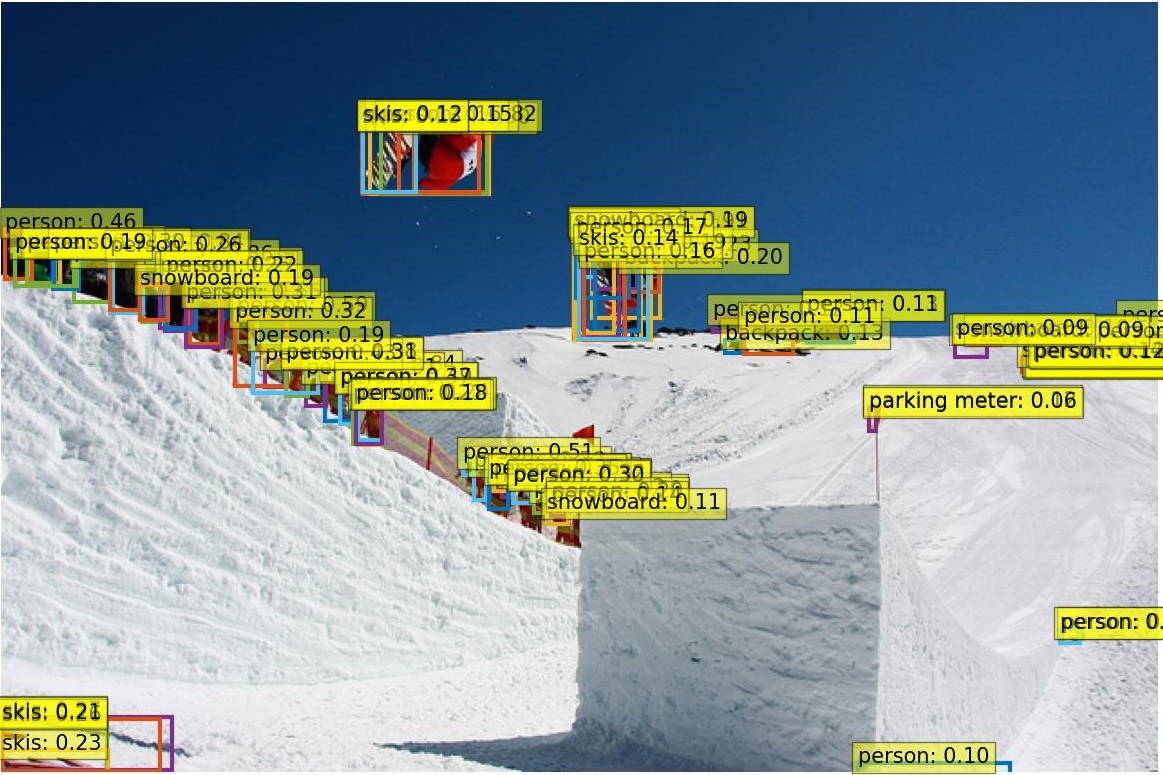} &\includegraphics[width=1.07\linewidth]{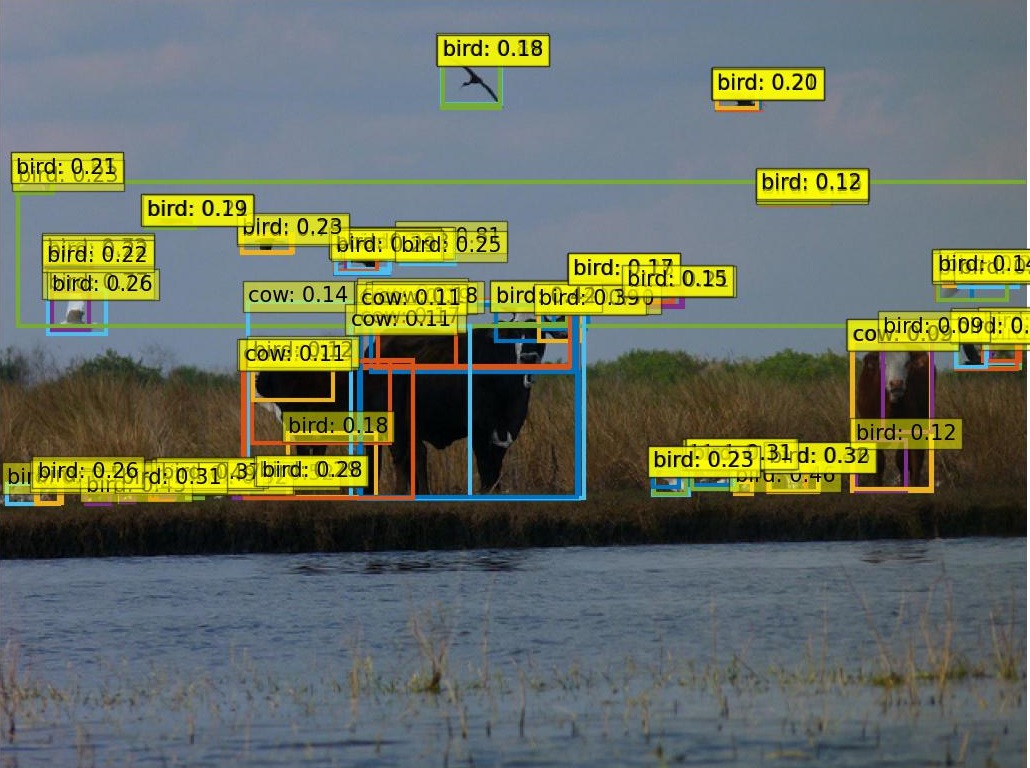}
		 & \includegraphics[width=1.07\linewidth]{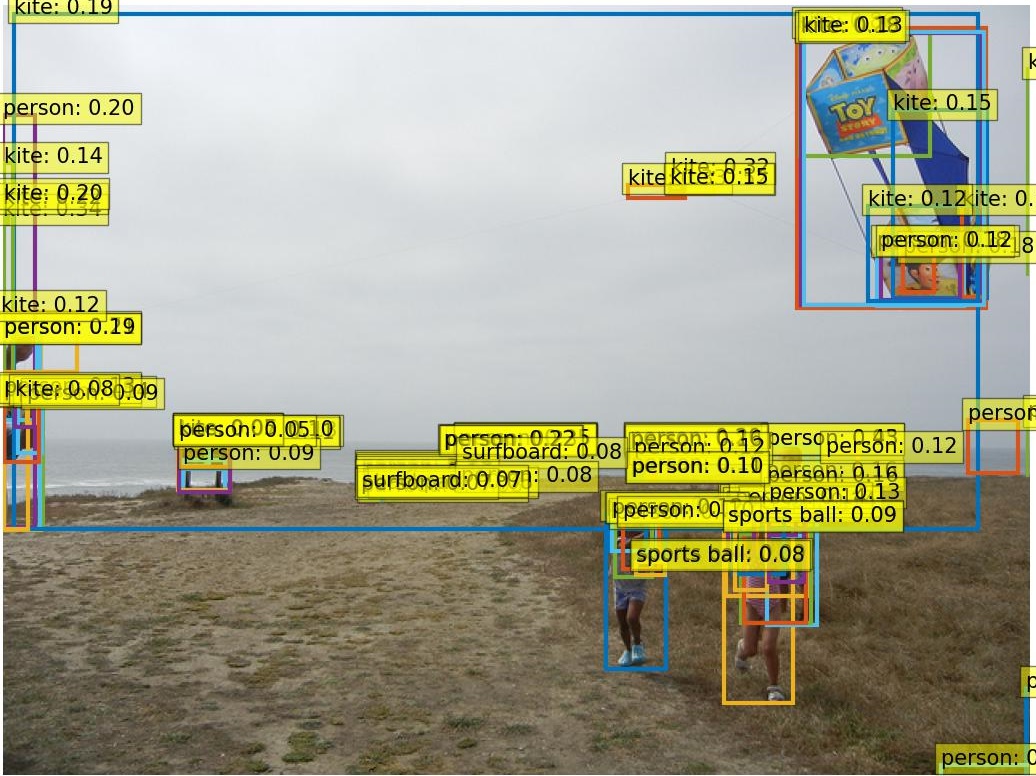}& \includegraphics[width=1.07\linewidth]{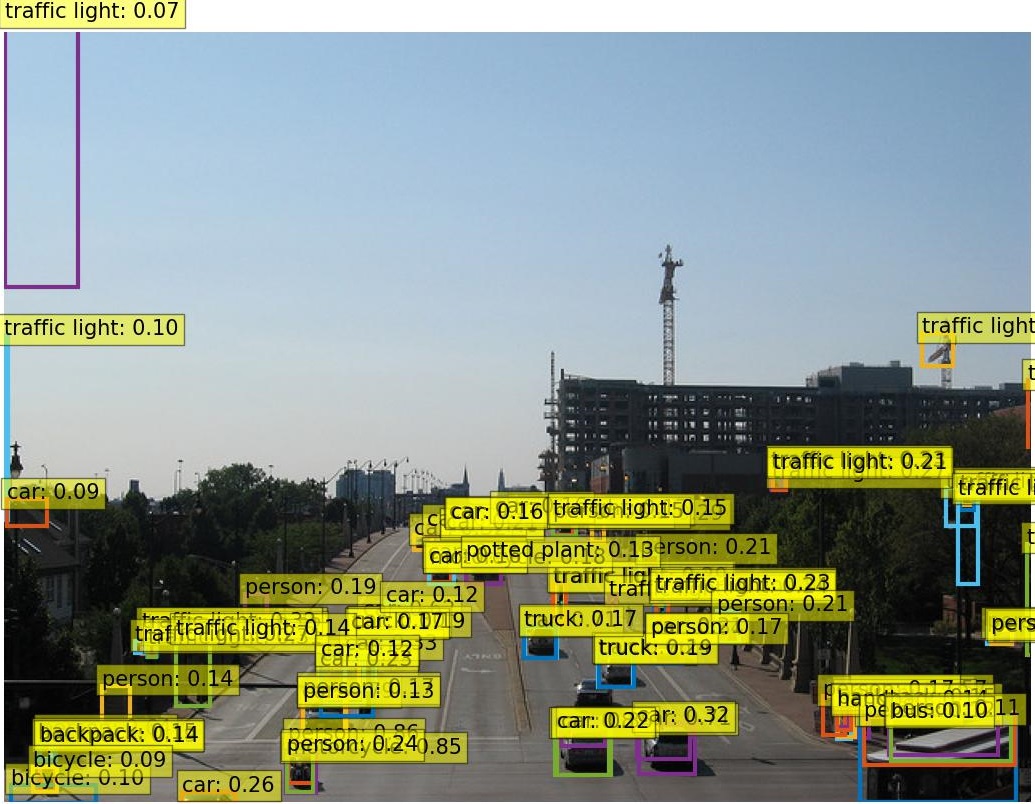}\\
      \hline

      	\end{tabular}}
   \caption{Examples of detection results on COCO dataset \cite{lin2014microsoft} for transformer-based SOTA small object detectors compared with Convolutional networks.}
\label{COCO}
\end{figure*}

In Table \ref{tab5}, we examine and evaluate the performance of all the techniques under review that have reported their results on MS COCO (compiled from their papers). The table provides information on the backbone architecture, GFLOPS/FPS (indicating the computational overhead and execution speed), number of parameters (indicating the scale of the model), mAP (mean average precision: a measure of object detection performance), and epochs (indicating the inference time and convergence properties). Additionally, a link to each method’s webpage is provided for further information. The methods are categorized into three groups: CNN-based, mixed, and transformer only methods. The top-performing methods for each metric are shown in the table’s last row. It should be noted that this comparison was only feasible for methods that have reported values for each specific metric. In instances where there is a tie, the method with the highest mean average precision was deemed the best. The default mAP values are for the "COCO 2017 val" set, while those for the  "COCO test-dev" set are marked with an asterisk. Please be aware that the reported mAP is only for objects with area$< 32^2$.
\begin{figure*}
	\ContinuedFloat\centering
	\scalebox{0.97}{\begin{tabular}{p{0.3cm} p{4.5cm} p{4cm} p{4cm} p{4cm}}
   \tiny \textbf{DETR}& \includegraphics[width=1.07\linewidth]{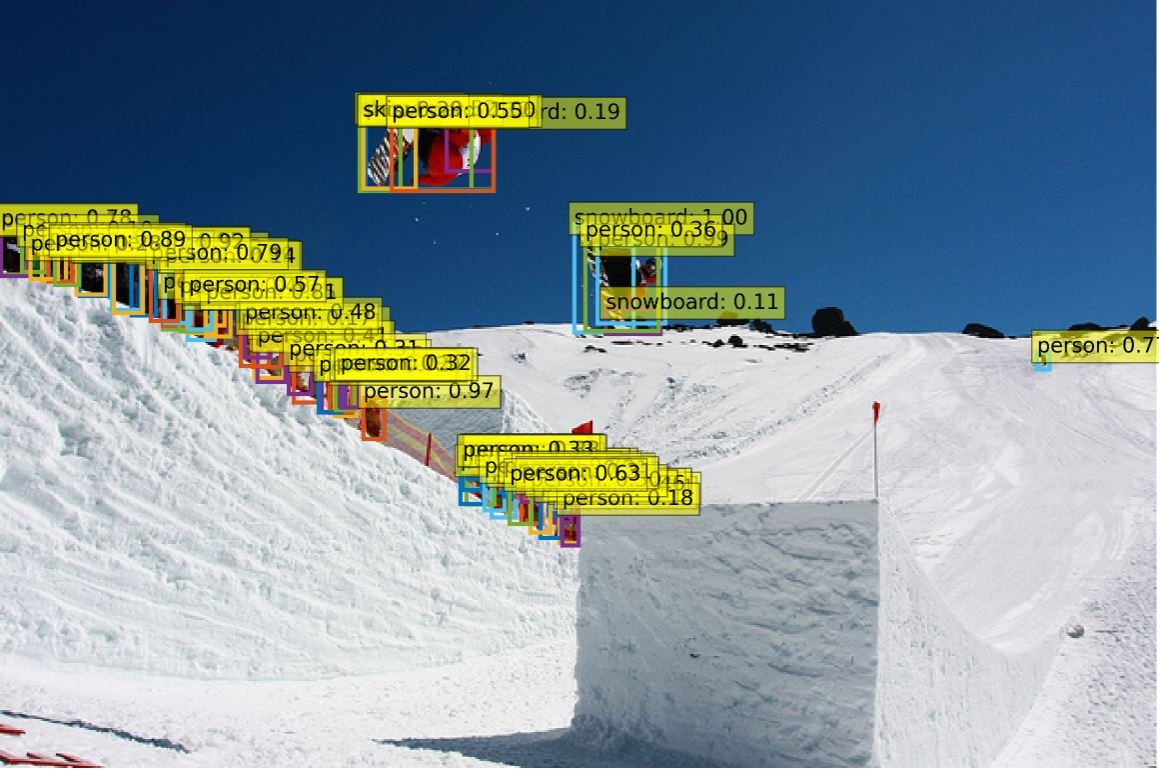} &\includegraphics[width=1.07\linewidth]{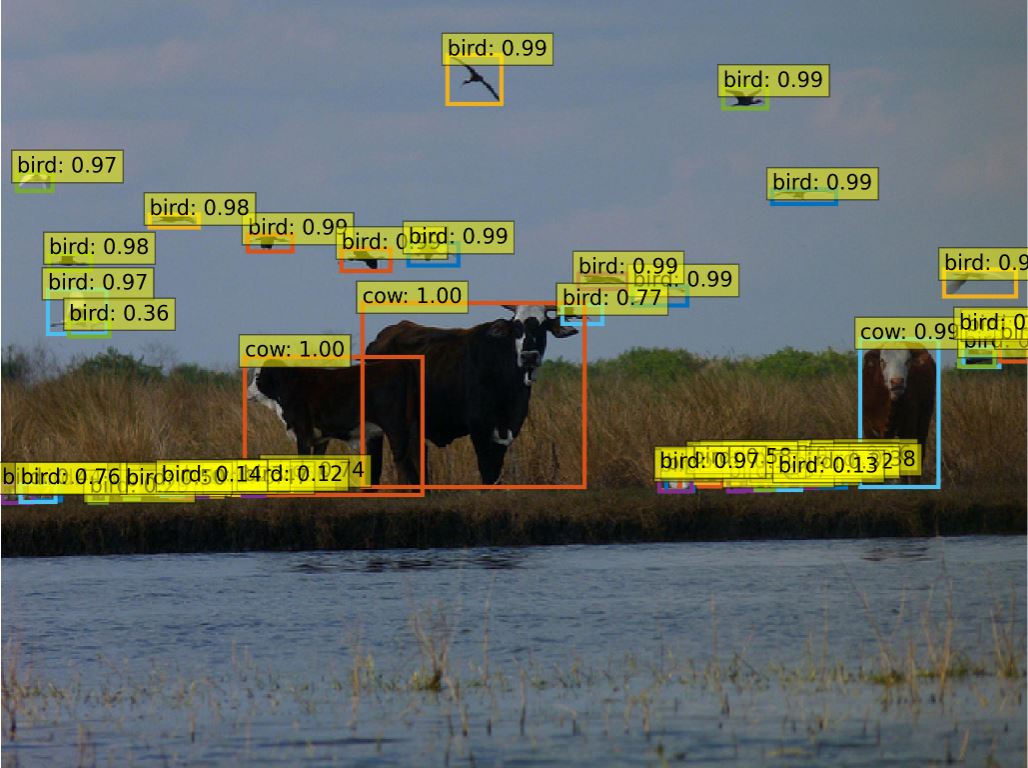}
		 & \includegraphics[width=1.07\linewidth]{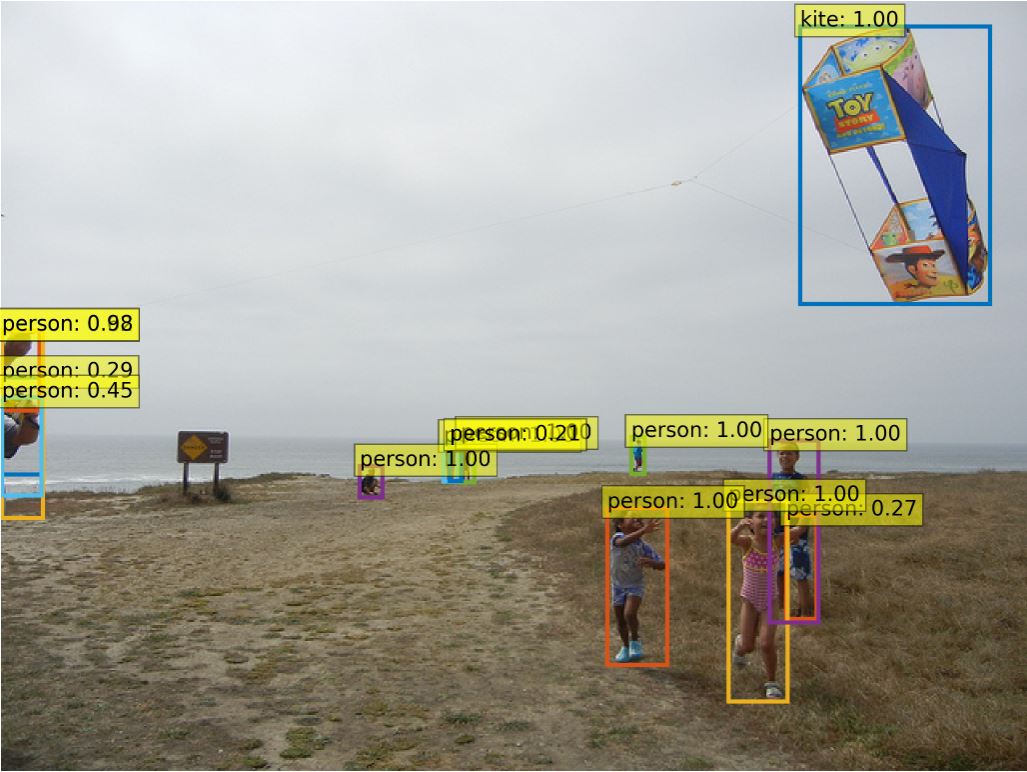}& \includegraphics[width=1.07\linewidth]{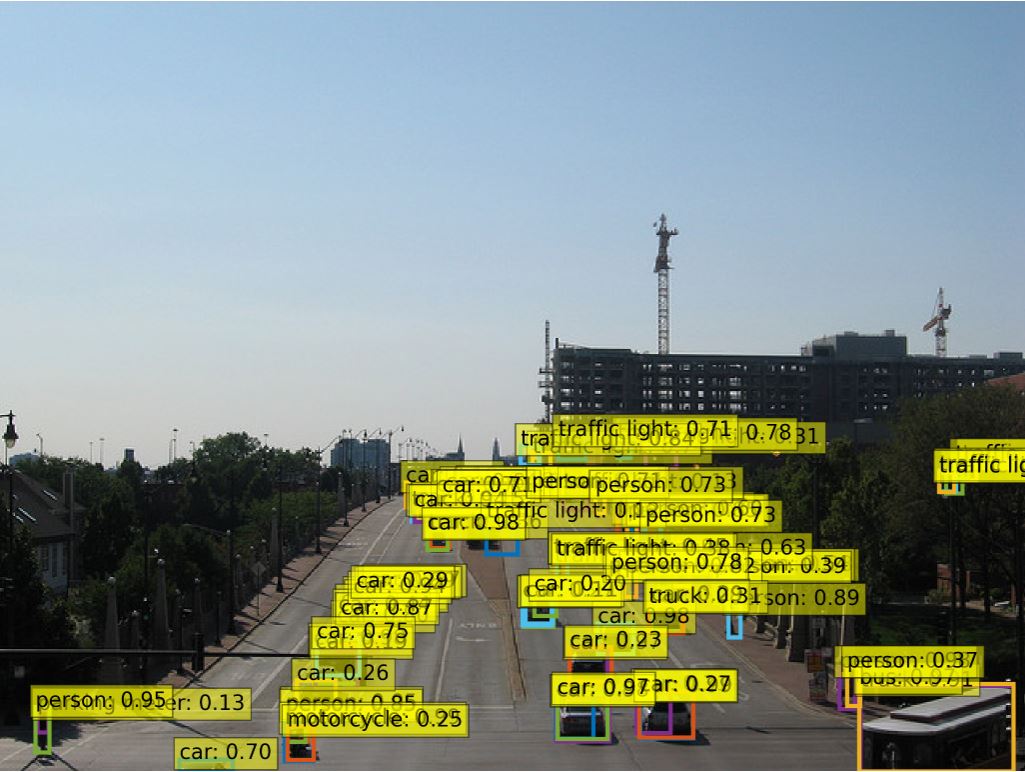}\\
\hline
\newpage
   \tiny \textbf{Faster RCNN}& \includegraphics[width=1.07\linewidth]{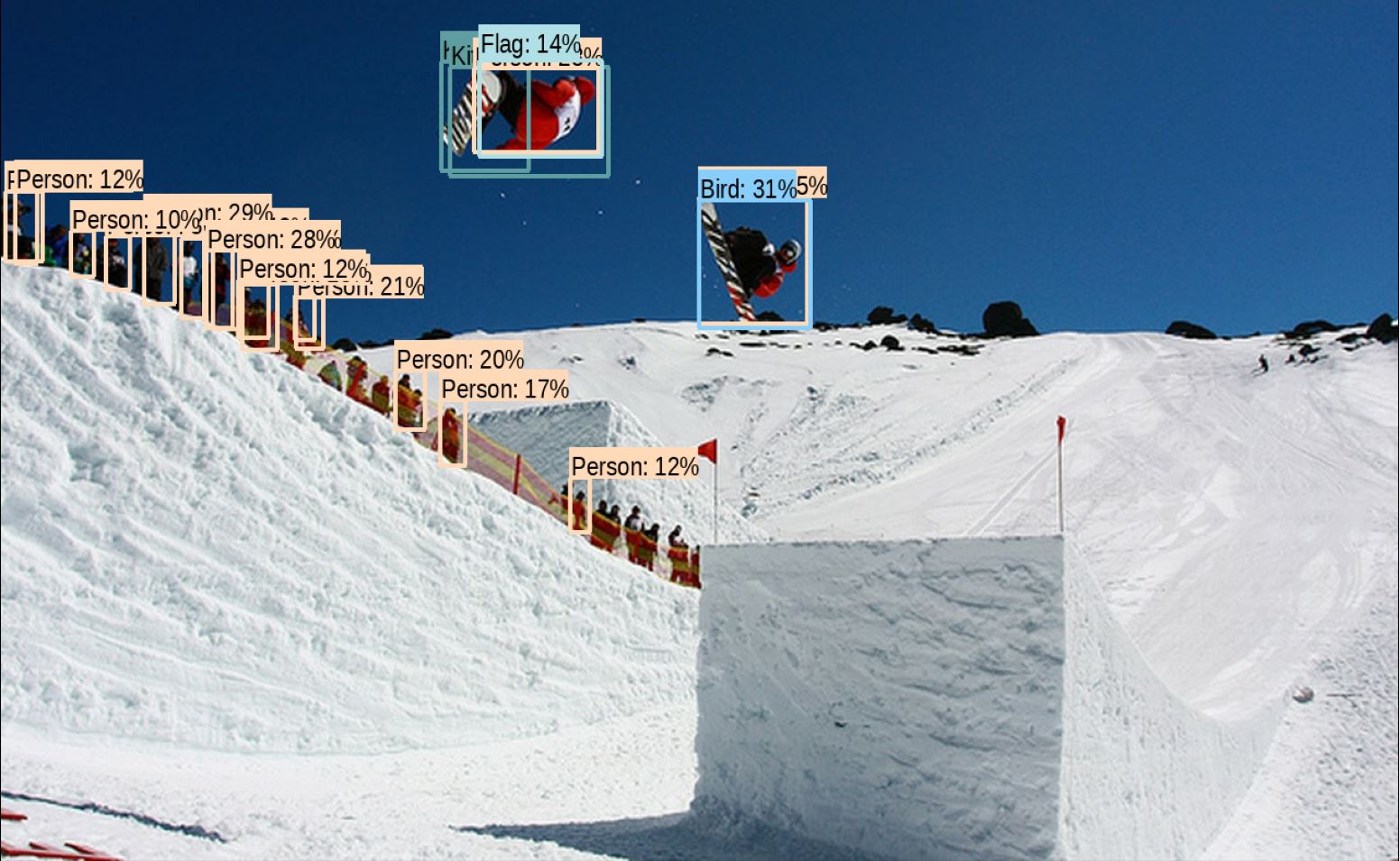} &\includegraphics[width=1.07\linewidth]{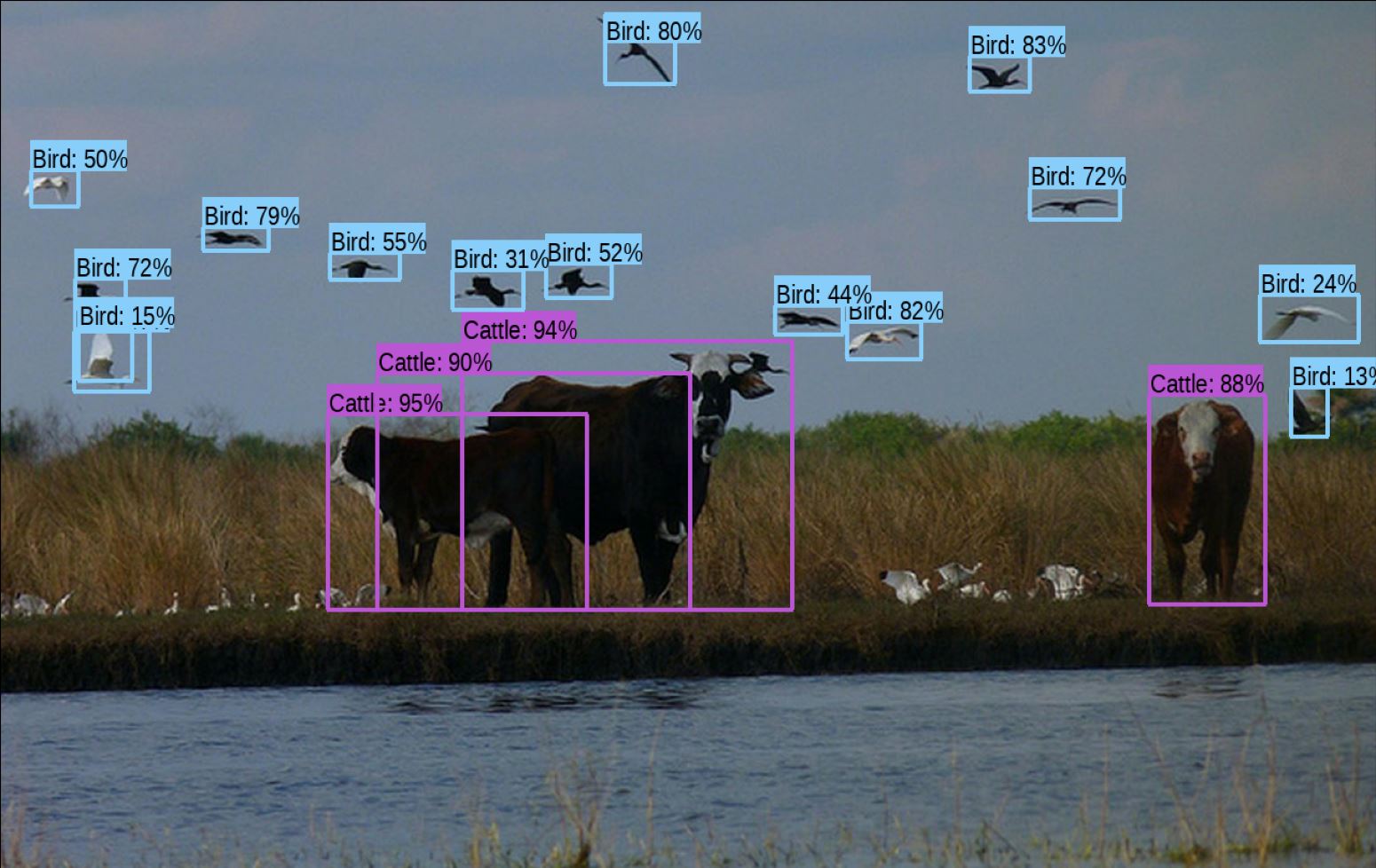}
		 & \includegraphics[width=1.07\linewidth]{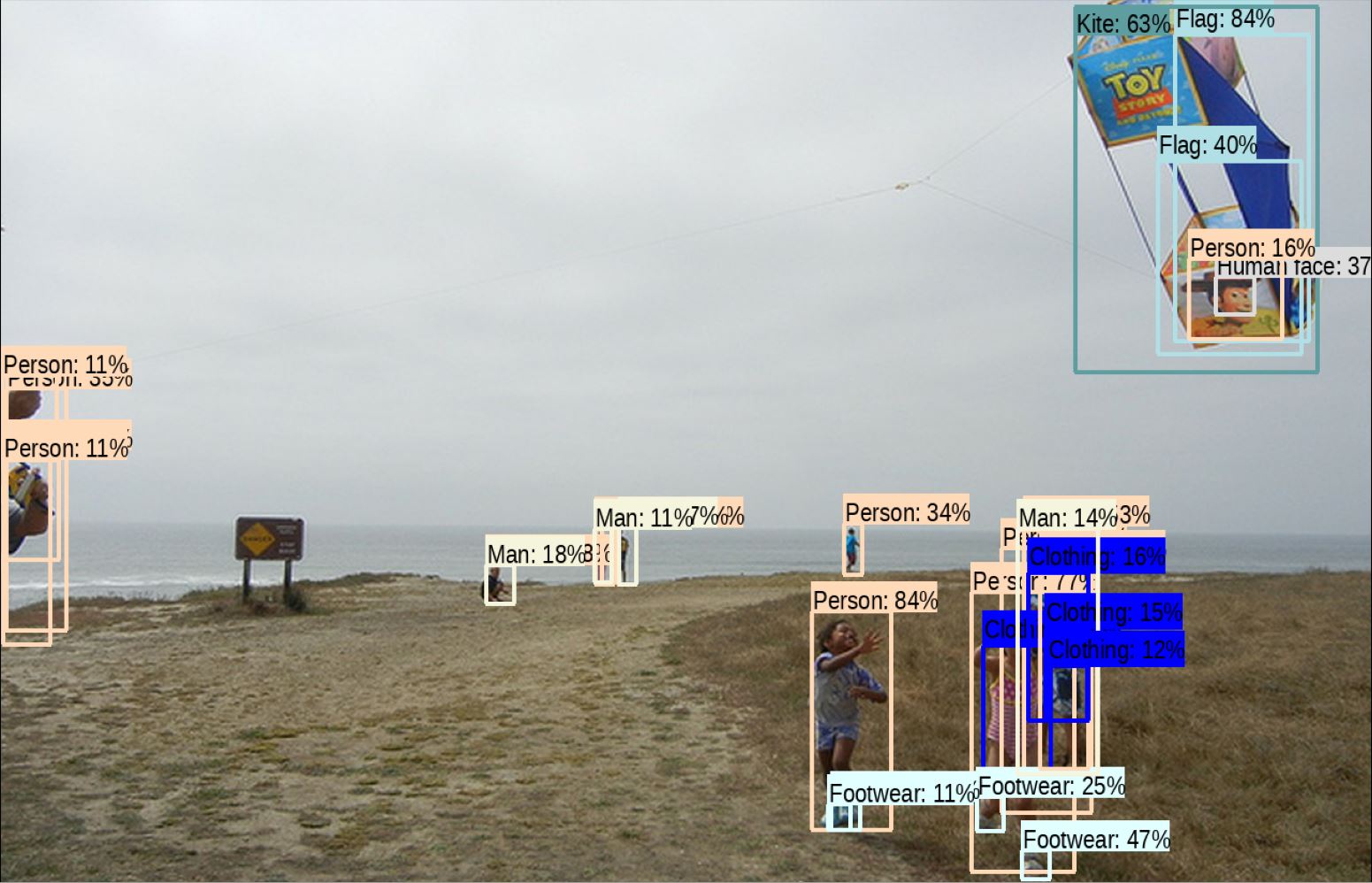}& \includegraphics[width=1.07\linewidth]{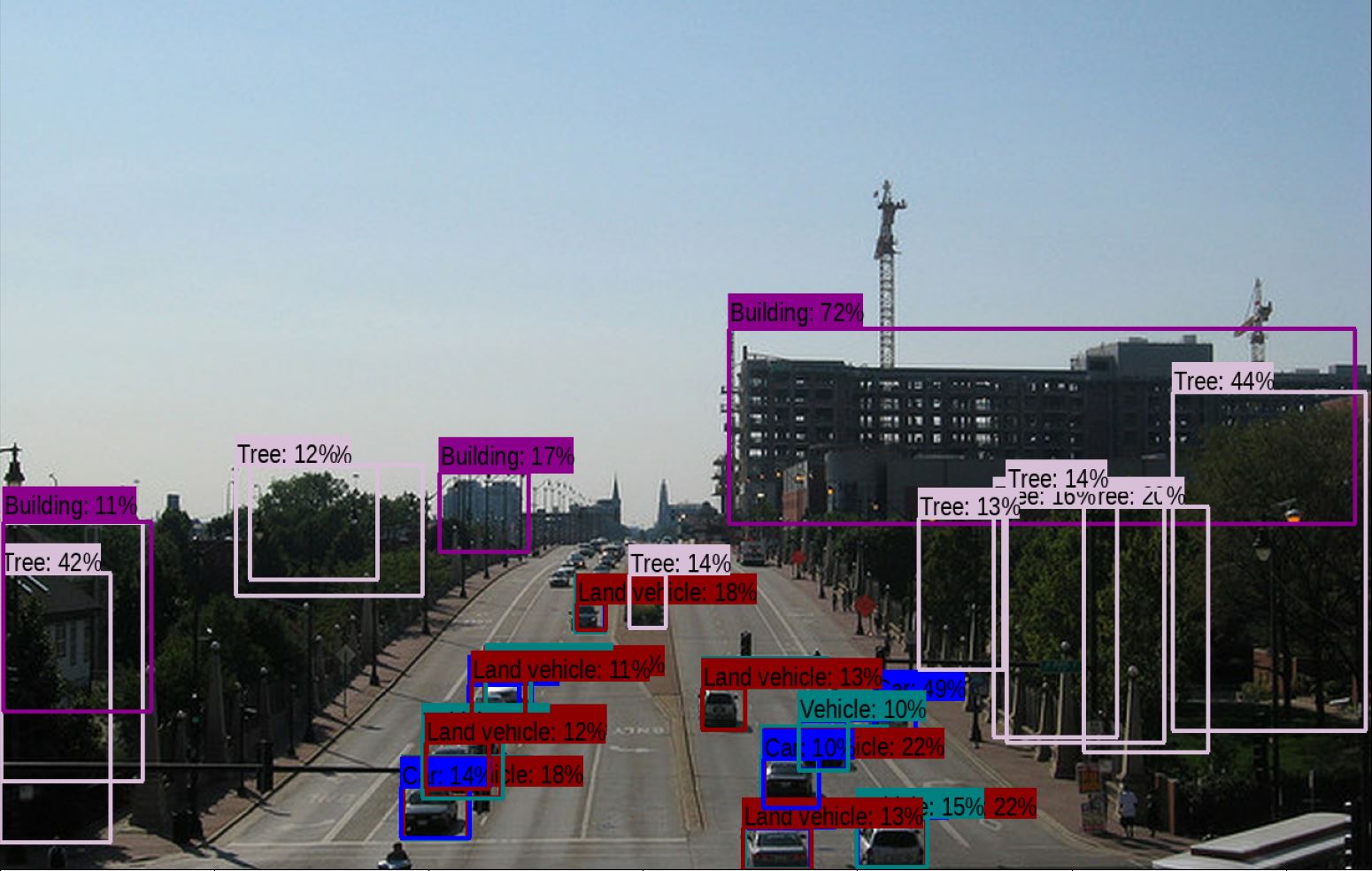}\\
         \hline
   \tiny \textbf{SSD}& \includegraphics[width=1.07\linewidth]{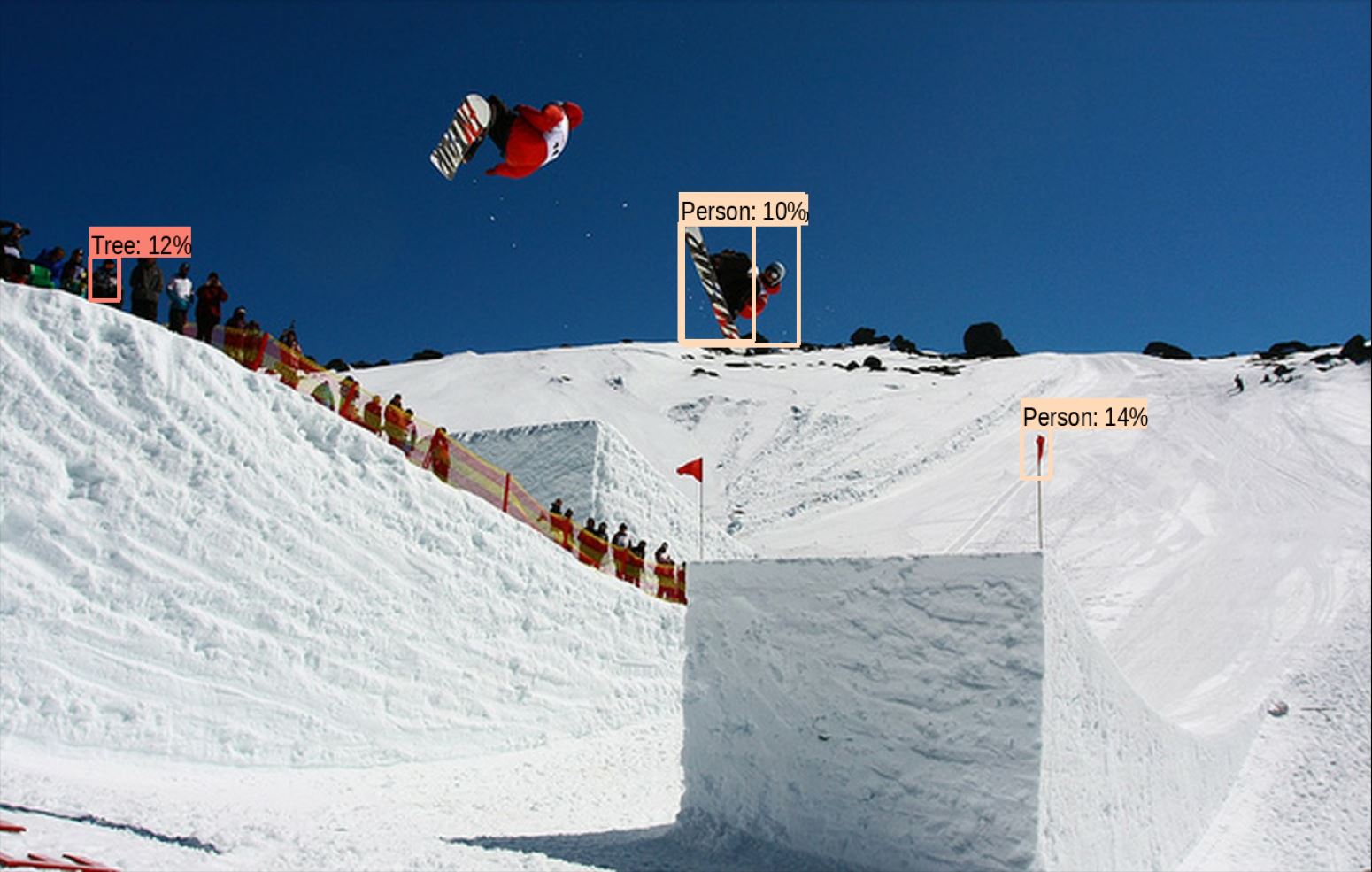} &\includegraphics[width=1.07\linewidth]{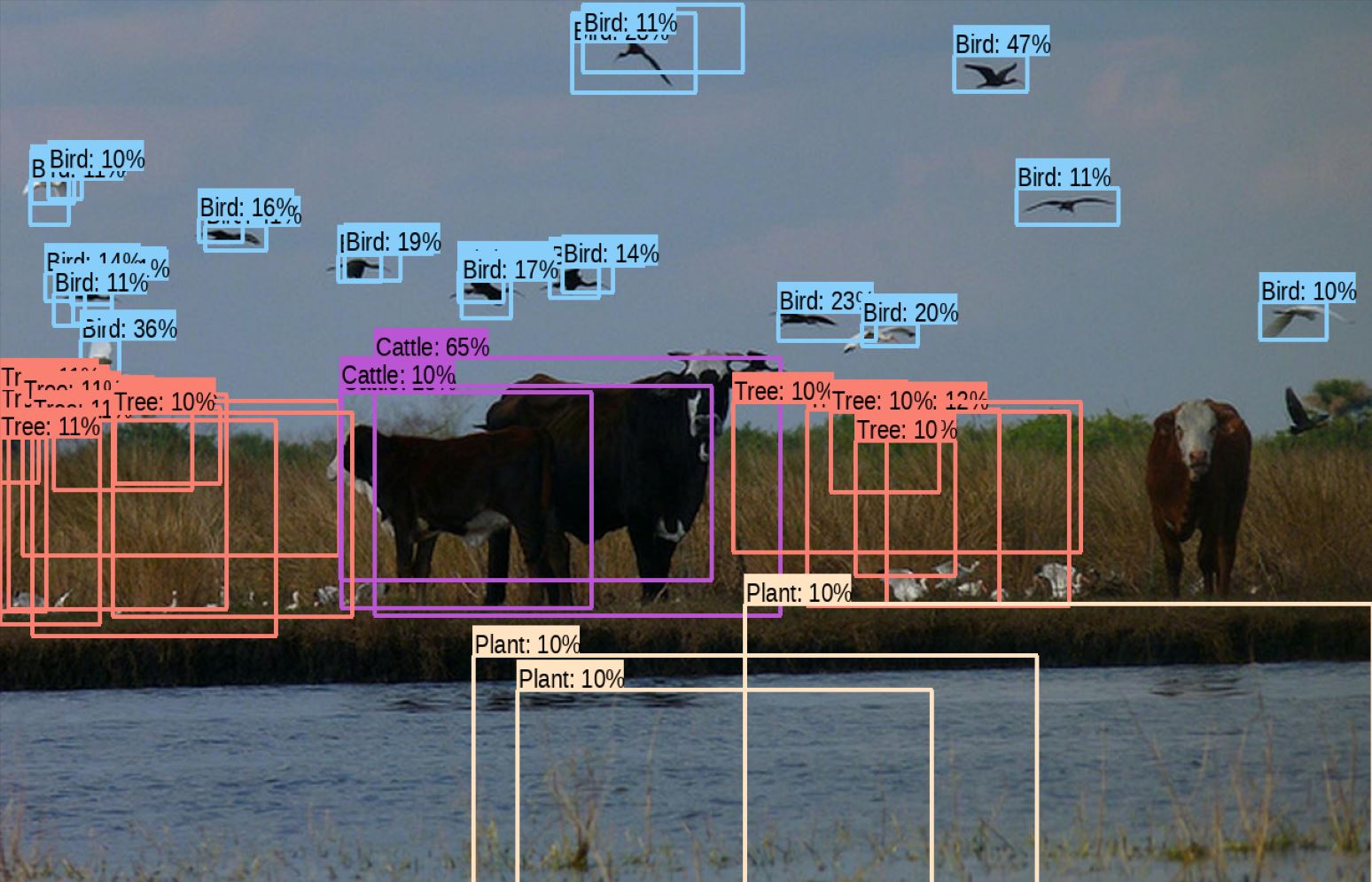}
		 & \includegraphics[width=1.07\linewidth]{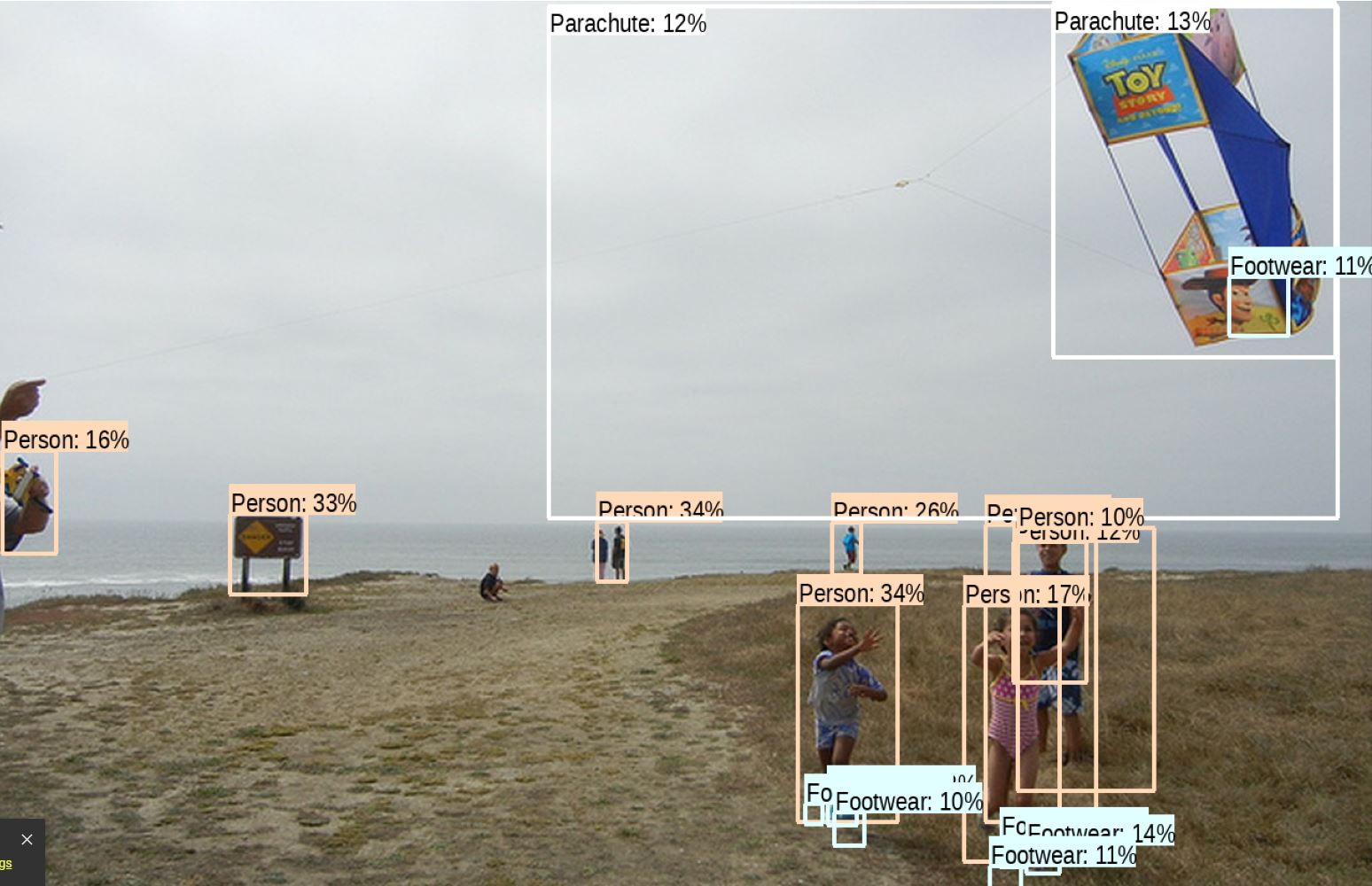}& \includegraphics[width=1.07\linewidth]{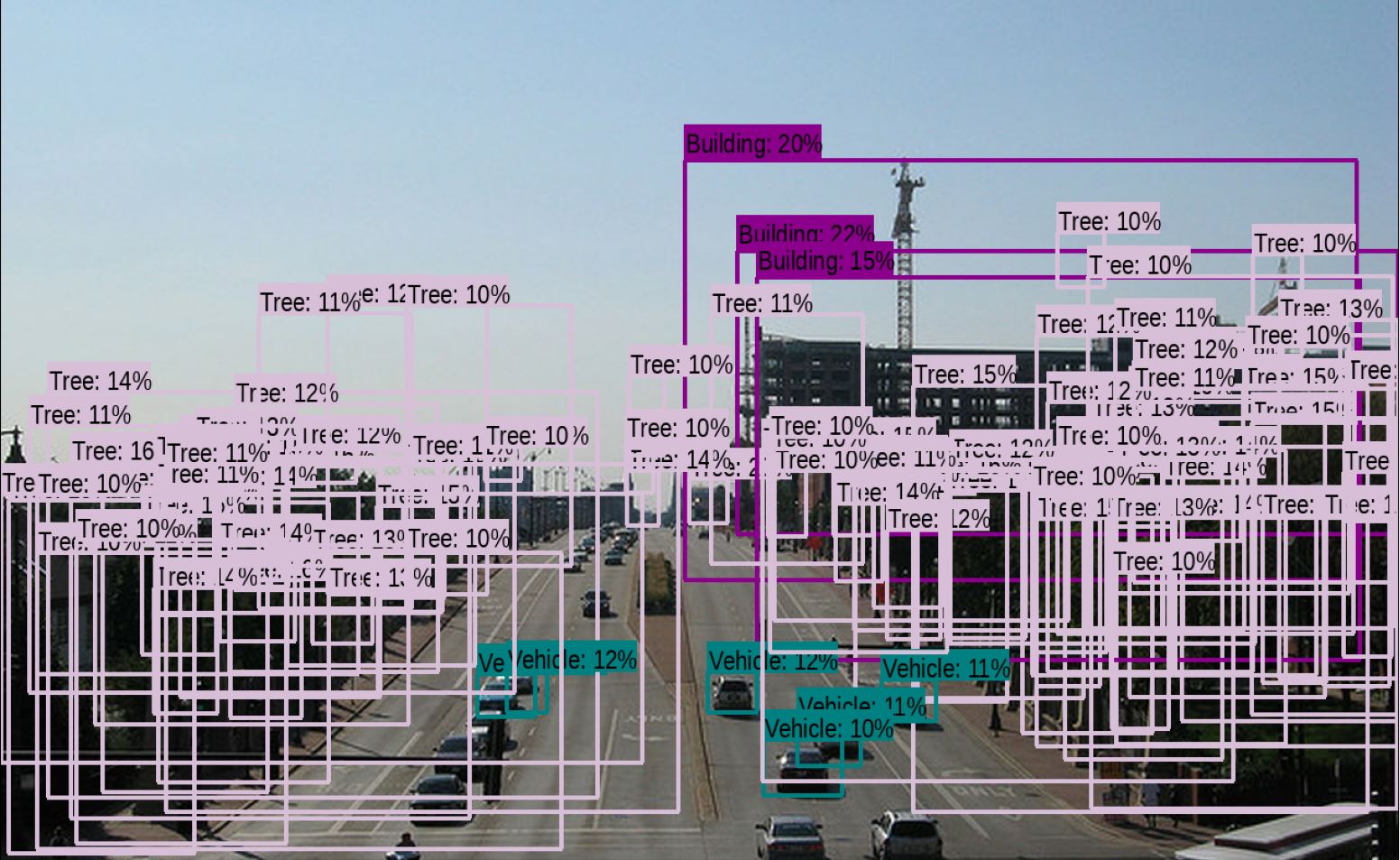}\\
   \hline
	\end{tabular}}
   \caption{Examples of detection results on COCO dataset \cite{lin2014microsoft} for transformer-based SOTA small object detectors compared with Convolutional networks.}
\end{figure*}

\begin{figure*}
	\centering
		   \includegraphics[width=0.25\linewidth]{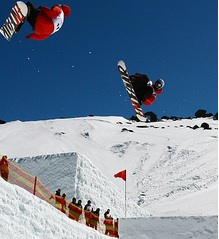} 
      \includegraphics[width=0.245\linewidth]{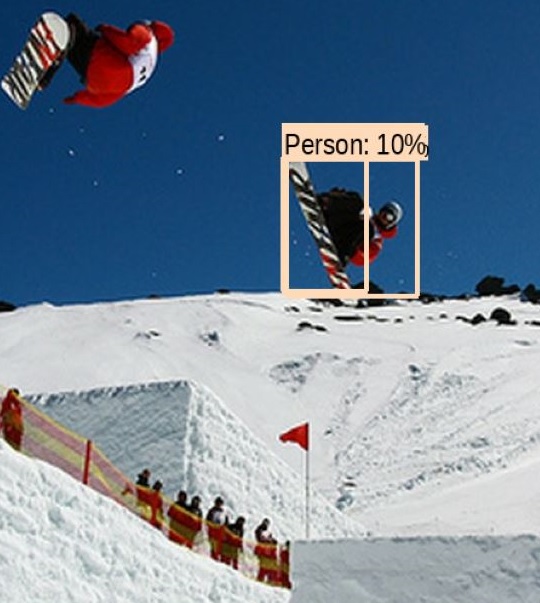}
     \includegraphics[width=0.235\linewidth]{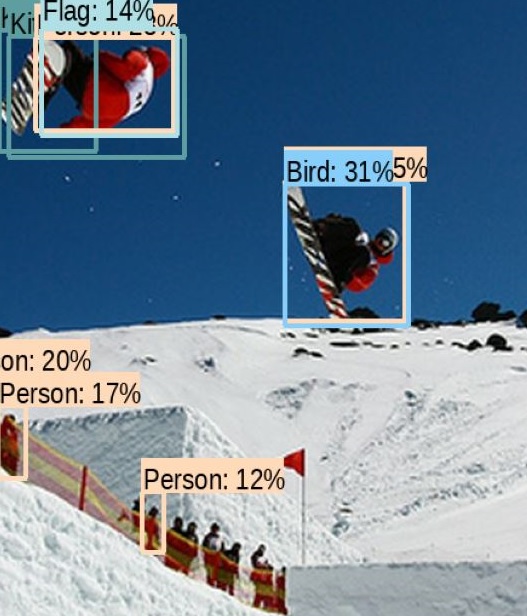}
		  \includegraphics[width=0.245\linewidth]{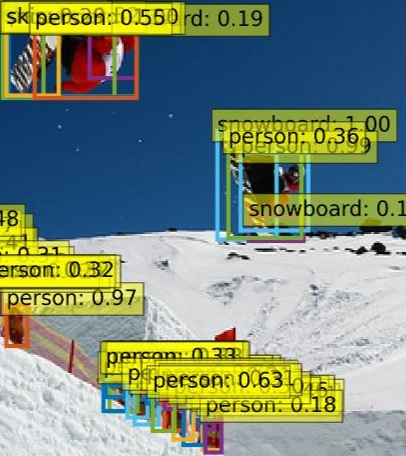} \\
     \includegraphics[width=0.24\linewidth]{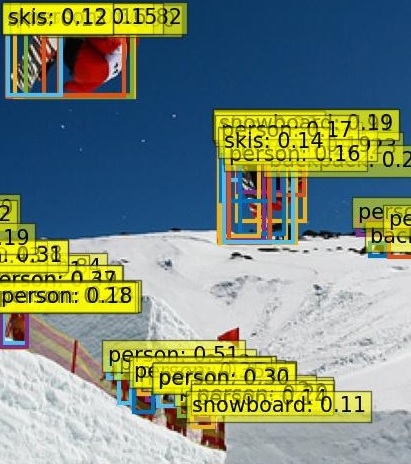} 
    \includegraphics[width=0.245\linewidth]{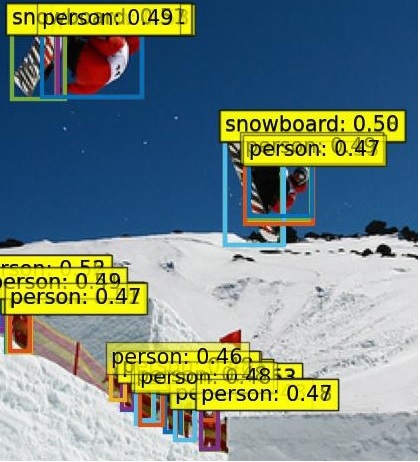} 
       \includegraphics[width=0.243\linewidth]{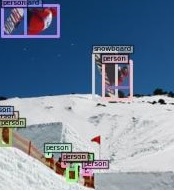}
   \includegraphics[width=0.245\linewidth]{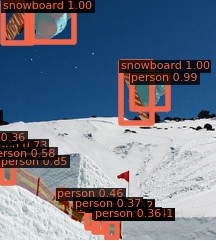}
	
   \caption{Detection results on a sample image when zoomed in. First row from the left: Input image, SSD, Faster RCNN, DETR. Second row from the left: ViDT, DETA-OB, DINO, CBNet v2.}
\label{COCOzoom}
\end{figure*}

Upon examining Table \ref{tab5}, it is obvious that most techniques benefit from using a mix of CNN and transformer architectures, essentially adopting hybrid strategies. Notably, Group DETR v2 which relies solely on a transformer-based architecture, attains a mAP of 48.4$\%$. However, achieving such a performance requires the adoption of additional techniques such as pre-training on two large-scale datasets and multi-scale learning. In terms of convergence, DINO outperforms by reaching stable results after just 12 epochs, while also securing a commendable mAP of 32.3$\%$. Conversely, the original DETR model has the fastest inference time and the lowest GFLOPS. FP-DETR stands out for having the lightest network with only 36M parameters.

Drawing from these findings, we conclude that pre-training and multi-scale learning emerge as the most effective strategies for excelling in small object detection. This may be attributed to the imbalance in downstream tasks and the lack of informative features in small objects.

Figure \ref{COCO} which spans two pages, along with its more detailed counterpart in Figure \ref{COCOzoom}, illustrates the detection results of various transformers and CNN-based methods. These are compared to each other using selected images from the COCO dataset and implemented by us using their public models available on their GitHub pages. The analysis reveals that Faster RCNN and SSD fall short in accurately detecting small objects. Specifically, SSD either misses most objects or generates numerous bounding boxes with false labels and poorly located bounding boxes. While Faster RCNN performs better, it still produces low-confidence bounding boxes and occasionally assigns incorrect labels.

In contrast, DETR has the tendency to over-estimate the number of objects, leading to multiple bounding boxes for individual objects. It is commonly noted t that DETR is prone to generating  false positives. Finally, among the methods evaluated, CBNet V2 stands out for its superior performance. As observed, it produces high confidence scores for the objects it detects, even though it may occasionally misidentify some objects.

\subsubsection{Small Object Detection in Aerial Images }
Another interesting use of detecting small objects is in the area of remote sensing. This field is particularly appealing because many organizations and research bodies aim to routinely monitor the Earth's surface through aerial images to collect both national and international data for statistics. While these images can be acquired using various modalities, this survey focuses only on non-SAR images. This is because SAR images have extensively been researched and deserve their own separate study. Nonetheless, the learning techniques discussed in this survey could also be applicable to SAR images.

\begin{figure}
	\centering
		   \includegraphics[width=1\linewidth]{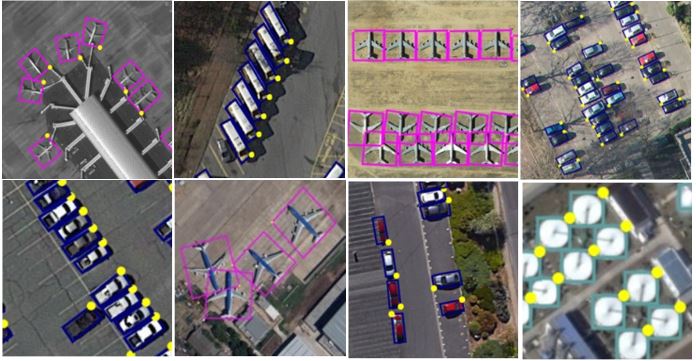} 
	
   \caption{Example of small objects in DOTA image dataset.}
\label{dota}
\end{figure}

In aerial images, objects often appear small due to their significant distance from the camera. The bird's-eye view also adds complexity to the task of object detection, as objects can be situated anywhere within the image. To assess the performance of transformer-based detectors designed for such applications, we selected the DOTA image dataset \cite{xia2018dota}, which has become a widely used benchmark in the field of object detection. Figure \ref{dota} displays some sample images from the DOTA dataset featuring small objects. The dataset includes a predefined Training set, Validation set, and Testing set. In comparison to generic applications, this particular application has received relatively less attention from transformer experts. However, as indicated in Table \ref{tab6} (results are compiled from papers), ReDet distinguishes itself throuh its multi-scale learning strategy and pre-training on the ImageNet dataset, achieving the highest precision value (80.89$\%$) and requiring only 12 training epochs. This mirrors the insights gained from the COCO dataset analysis, suggesting that optimal performance can be attained by addressing imbalances in downstream tasks and including informative features from small objects.

\begin{table*}[]
\caption{Detection performance (\%) for small-scale objects on DOTA image dataset \cite{xia2018dota}. The top section shows results for CNN-based techniques, the middle section shows results for mixed architectures. MS: Multi-scale network, FT: Fine-tuned, FPN: Feature pyramid network, IN: Pre-trained on ImageNet.}
\centering
\label{tab6}
\scalebox{1}{\begin{tabular}{lcccccc}
\toprule
Model& Backbone& FPS $\uparrow$ & $\#$params$\downarrow$ &$\text{mAP}$  $\uparrow$& Epochs$\downarrow$&URL  \\ 
\hline

Rotated Faster RCNN-MS~(NeurIPS2015)\cite{ren2015faster}& ResNet101&-- & 64M&67.71& 50&\href{https://github.com/open-mmlab/mmrotate/tree/main/configs/rotated_faster_rcnn}{\textcolor{blue}{Link}}\\

SSD~(ECCV2016) \cite{liu2016ssd} &-- &-- & --&56.1& --& \href{https://github.com/pierluigiferrari/ssd_keras}{\textcolor{blue}{Link}}\\ 
RetinaNet-MS~(ICCV2017)\cite{lin2017focal}& ResNet101& --&\textbf{59M}&66.53&50&\href{https://github.com/DetectionTeamUCAS/RetinaNet_Tensorflow}{\textcolor{blue}{Link}}\\

ROI-Transformer-MS-IN~(CVPR2019) \cite{ding2019learning, wang2023aerial} & ResNet50&-- & --&80.06&\textbf{12} & \href{https://github.com/open-mmlab/mmrotate/blob/main/configs/roi_trans/README.md}{\textcolor{blue}{Link}}\\

Yolov5~(2020)\cite{jocher2020yolov5}&-- &\textbf{95}& --&64.5& --& \href{https://github.com/ultralytics/yolov5}{\textcolor{blue}{Link}}\\

ReDet-MS-FPN~(CVPR2021)\cite{han2021redet}& ResNet50& --&--&80.1&--&\href{https://github.com/csuhan/ReDet}{\textcolor{blue}{Link}}\\

\hline

O$^2$DETR-MS~(arXiv2021)\cite{ma2021oriented}& ResNet101&-- & 63M&70.02&50&--\\

O$^2$DETR-MS-FT~(arXiv2021)\cite{ma2021oriented}& ResNet101&-- & --& 76.23&62&--\\

O$^2$DETR-MS-FPN-FT~(arXiv2021)\cite{ma2021oriented}& ResNet50&-- & --& 79.66&--&--\\

SPH-Yolov5~(RS2022) \cite{gong2022swin} & Swin Transformer-base &51 &-- & 71.6& 150&--\\

AO2-DETR-MS~(TCSVT2022)\cite{dai2022ao2}& ResNet50& --& --& 79.22&--&\href{https://github.com/Ixiaohuihuihui/AO2-DETR}{\textcolor{blue}{Link}}\\

MDCT~(RS2023)\cite{chen2023mdct}&-- &-- & --&75.7& --& --\\

ReDet-MS-IN~(arXiv2023)\cite{wang2023aerial}& ViTDet, ViT-B&--&--&\textbf{80.89}&\textbf{12}&\href{https://github.com/csuhan/ReDet}{\textcolor{blue}{Link}}\\

\hline
Best Results&NA&Yolov5&RetinaNet&ReDet-MS-IN&ReDet-MS-IN&NA\\
\bottomrule
\end{tabular}}
\end{table*}

\subsubsection{Small Object Detection in Medical Images}
\begin{figure*}
	\centering
		   \includegraphics[width=0.98\linewidth]{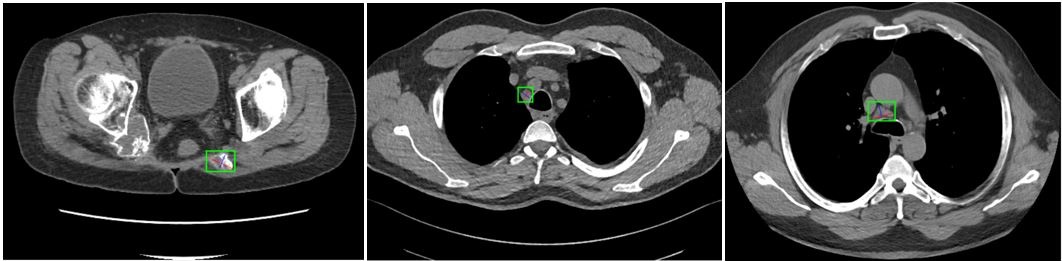} 
	
   \caption{Example of small abnormalities in DeepLesion image dataset \cite{yan2018deeplesion}.}
\label{Lesion}
\end{figure*}

In the field of medical imaging, specialists are often tasked with early detection and identification of anomalies. Missing even barely visible or small abnormal cells can lead to serious repercussions for patients,  including cancer and life-threatening conditions. These small-sized objects can be found as abnormalities in the retina of diabetic patients, early tumors, vascular plaques, etc. Despite the critical nature and potential life-threatening impact of this research area, only a handful of studies have tackled the challenges associated with detecting small objects in this crucial application. For those interested in this topic, the DeepLesion CT image dataset \cite{yan2018deeplesion} has been selected as the benchmark due to the availability of the results for this particular dataset \cite{li2021systematic}. Sample images from this dataset are shown in Figure \ref{Lesion}. This dataset is divided into three sets: training (70$\%$), validation (15$\%$), and test (15$\%$) sets \cite{shou2022object}.   Table \ref{tab7} compares the accuracy and mAP of three transformer-based studies against both two-stage and one-stage detectors (results are compiled from their papers). The MS Transformer emerges as the best technique with this dataset, albeit with limited competition. Its primary innovation lies in self-supervised learning and the incorporation of a masking mechanism within a hierarchical transformer model. Overall, with an accuracy of 90.3$\%$ and an mAP of 89.6$\%$, this dataset appears to be less challenging compared to other medical imaging tasks, especially considering that some tumor detection tasks are virtually invisible to the human eyes.
\begin{table}[]
\caption{Detection performance (\%) for DeepLesion CT image dataset \cite{yan2018deeplesion}. The top section shows results for CNN-based techniques, the middle section shows results for mixed architectures. }
\centering
\label{tab7}
\scalebox{1}{\begin{tabular}{ccc}
\toprule
Model& Accuracy $\uparrow$  &$\text{mAP}^{0.5}$  $\uparrow$\\ 
\hline

Faster RCNN~(NeurIPS2015)\cite{ren2015faster}& 83.3 &83.3\\ 

Yolov5 \cite{jocher2020yolov5}&85.2 &88.2\\ 

\hline

DETR~(ECCV2020)\cite{carion2020end}&86.7 &87.8\\

Swin Transformer& 82.9&81.2\\

MS Transformer~(CIN2022)\cite{shou2022object}&\textbf{90.3}&\textbf{89.6}\\
\hline

Best Results&MS Transformer&MS Transformer\\
\bottomrule
\end{tabular}}
\end{table}

\begin{figure*}
\begin{center}
   \includegraphics[width=0.98\linewidth]{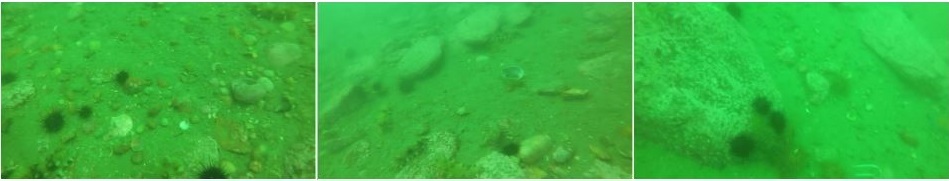}
\end{center}
   \caption{Examples of low quality images in URPC2018 image dataset.}
\label{URPC}
\end{figure*}

\begin{figure*}
\begin{center}
   \includegraphics[width=1\linewidth]{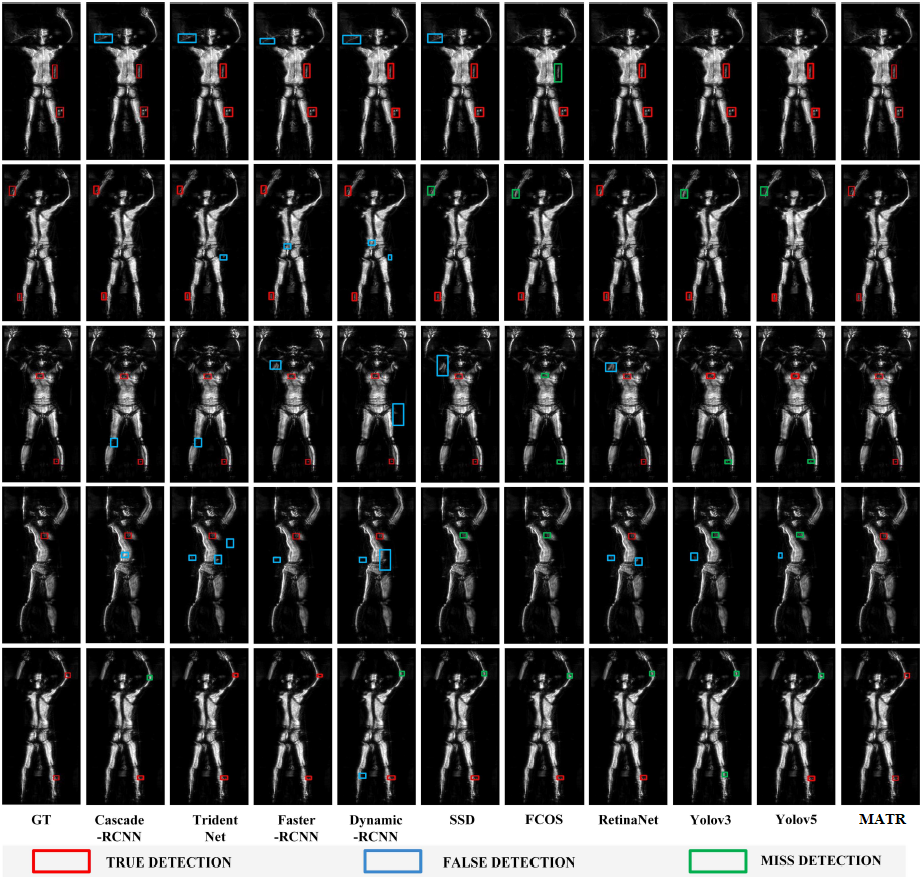}
\end{center}
   \caption{Examples of detection results on AMMW image dataset \cite{liu2019concealed} for SOTA small object detectors (figure from \cite{sun2022multi}).}
\label{figawmm}
\end{figure*}

\subsubsection{Small Object Detection in Underwater Images}
With the growth of underwater activities, the demand to monitor hazy and low-light environments has increased for purposes like ecological surveillance, equipment maintenance, and monitoring of wreck fishing. Factors like scattering and light absorption of the water, make the SOD task even more challenging. Example images of such challenging environments are displayed in Figure \ref{URPC}. Transformer-based detection methods should not only be adept at identifying small objects but also need to be robust against the poor image quality found in deep waters, as well as variations in color channels due to differing rates of light attenuation for each channel.

Table \ref{tab10}  shows the performance metrics reported in existing studies for this dataset (results are compiled from their papers). HTDet is the sole transformer-based technique identified for this specific application. It significantly outperforms the SOTA CNN-based method by a huge margin (3.4$\%$ in mAP). However, the relatively low mAP scores confirm that object detection in underwater images remains a difficult task. It is worth noting that the training set of the URPC 2018 contains 2901 labeled images, and the testing set contains 800 unlabeled images \cite{chen2023htdet}.

\begin{table}[]
\caption{Detection performance (\%) for URPC2018 dataset \cite{lin2020roimix}. The top section shows results for CNN-based techniques, the middle section shows results for mixed architectures. }
\centering
\label{tab10}
\scalebox{0.85}{\begin{tabular}{lccc}
\toprule
Model& $\#$Params$\downarrow$&  $\text{mAP}^{@[0.5,0.95]}$  $\uparrow$ & $\text{mAP}^{0.5}$  $\uparrow$ \\ 
\hline

Faster RCNN~(NeurIPS2015)\cite{ren2015faster}&33.6M& 16.4&--\\ 

Cascade RCNN~(CVPR2018)\cite{cai2018cascade}& 68.9M& 16&--\\

Dynamic RCNN~(ECCV2020) \cite{zhang2020dynamic}&41.5M& 13.3&-- \\

Yolov3 \cite{jocher2020yolov5}& 61.5M& 19.4&--\\ 

RoIMix~(ICASSP2020) \cite{lin2020roimix}& --&-- &\textbf{74.92} \\
\hline

HTDet~(RS2023) \cite{chen2023htdet}&\textbf{7.7M}&\textbf{22.8}&--\\
\hline
Best Results&HTDet&HTDet&RoIMix\\
\bottomrule
\end{tabular}}
\end{table}

\subsubsection{Small Object Detection in Active Milli-Meter Wave Images}
Small objects can easily be concealed or hidden from normal RGB cameras, for example, within a person's clothing at an airport. Therefore, active imaging techniques are essential for security purposes. In these scenarios, multiple images are often captured from different angles to enhance the likelihood of detecting even minuscule objects. Interestingly, much like in the field of medical imaging, transformers are rarely used for this particular application.

In our study, we focused on the detection performance of existing techniques using the AMMW Dataset \cite{liu2019concealed} as shown in Table \ref{tab9} (results are compiled from their papers). We have identified that MATR emerged as the sole technique that combines transformer and CNNs for this dataset. Despite being the only transformer-based technique, it could significantly improve the SOD performance (5.49$\%\uparrow$ in $\text{mAP}^{0.5}$ with respect to Yolov5 and 4.22 $\%\uparrow$ in $\text{mAP}^{@[0.5,0.95]}$ with respect to TridentNet) with the same backbone (ResNet50). Figure \ref{figawmm} visually compares MATR with other SOTA CNN-based techniques. Combining images from different angles largely helps to identify even small objects within this imaging approach. For training and testing, 35426  and  4019 images were used, respectively \cite{sun2022multi}.

\begin{table}[]
\caption{Detection performance (\%) for AMWW image dataset \cite{liu2019concealed}. The top section shows results for CNN-based techniques, the middle section shows results for mixed architectures. }
\centering
\label{tab9}
\scalebox{0.87}{\begin{tabular}{lccc}
\toprule
Model& Backbone&  $\text{mAP}^{0.5}$  $\uparrow$& $\text{mAP}^{@[0.5,0.95]}$  $\uparrow$ \\ 
\hline

Faster RCNN~(NeurIPS2015)\cite{ren2015faster}&ResNet50 &70.7& 26.83\\ 

Cascade RCNN~(CVPR2018)\cite{cai2018cascade}&ResNet50 & 74.7 & 27.8\\

TridentNet~(ICCV2019) \cite{li2019scale}&ResNet50&77.3&29.2\\

Dynamic RCNN~(ECCV2020) \cite{zhang2020dynamic}& ResNet50& 76.3& 27.6\\

Yolov5 \cite{jocher2020yolov5}&ResNet50 &76.67& 28.48\\

\hline

MATR~(TCSVT2022) \cite{sun2022multi}& ResNet50& \textbf{82.16} & \textbf{33.42}\\
\hline
Best Results&NA&MATR&MATR\\
\bottomrule
\end{tabular}}
\end{table}

\begin{table}[]
\caption{Detection performance (\%) for ImageNet VID dataset \cite{russakovsky2015imagenet} for small objects. The top section shows results for CNN-based techniques, the middle section shows results for mixed architectures. PT: Pre-trained on MS COCO. }
\centering
\label{tab11}
\scalebox{0.83}{\begin{tabular}{lcc}
\toprule
Model&  Backbone& $\text{mAP}^{@[0.5,0.95]}$  $\uparrow$  \\ 
\hline

Faster RCNN~(NeurIPS2015)\cite{ren2015faster}+SELSA\cite{wu2019sequence}& ResNet50& 8.5 \\ 

\hline

Deformable-DETR-PT \cite{zhu2021deformable}&ResNet50&10.5\\

Deformable-DETR\cite{zhu2021deformable}+TransVOD-PT\cite{zhou2022transvod}&ResNet50&11\\

DAB-DETR\cite{liu2022dab}+FAQ-PT\cite{cui2023faq}&ResNet50&12\\

Deformable-DETR\cite{zhu2021deformable}+FAQ-PT\cite{cui2023faq}&ResNet50&\textbf{13.2}\\
\hline

Best Results&NA&Deformable-DETR+FAQ\\
\bottomrule
\end{tabular}}
\end{table}

\subsubsection{Small Object Detection in Videos}
The field of object detection in videos gained considerable attention recently, as the temporal information in videos can improve the detection performance.  To benchmark the SOTA techniques, the ImageNet VID dataset has been used with results specifically focused on the dataset's  small objects. This dataset includes 3862 training videos and 555 validation videos with 30 classes of objects.  Table \ref{tab11} reports the mAP of several recently developed transformer-based techniques (results are compiled from their papers). While transformers are increasingly being used in video object detection, their performance in SOD remains less explored. Among the methods that have reported SOD performance on the ImageNet VID dataset, Deformable DETR with FAQ stands out for achieving the highest performance- although it is notably low at 13.2 $\%$ for $\text{mAP}^{@[0.5,0.95]}$). This highlights a significant research gap in the area of video-based SOD.

\section{Discussion}\label{findings}

In this survey article, we explored how transformer-based approaches can address the challenges of SOD. Our taxonomy divides transformer-based small object detectors into seven main categories: object representation, fast attention (useful for high-resolution and multi-scale feature maps), architecture and block modification, spatio-temporal information, improved feature representation, auxiliary techniques, and fully transformer-based detectors.

When juxtaposing this taxonomy with the one for CNN-based techniques \cite{rekavandi2022guide}, we observe that some of these categories overlap, while others are unique to transformer-based techniques. Certain strategies are implicitly embedded into transformers, such as attention and context learning, which are performed via the self and cross-attention modules in the encoder and decoder. On the other hand, multi-scale learning, auxiliary tasks, architecture modification, and data augmentation are commonly used in both paradigms. However, it is important to note that while CNNs handle spatio-temporal analysis through 3D-CNN, RNN, or feature aggregation over time, transformers achieve this by using successive spatial and temporal transformers or by updating object queries for successive frames in the decoder.

We have observed that pre-training and multi-scale learning stand out as the most commonly adopted strategies, contributing to state-of-the-art performance across various datasets performance on different datasets. Data fusion is another approach widely used for SOD. In the context of video-based detection systems, the focus is on effective methods for collecting temporal data and integrating it into the frame-specific detection module.

While transformers have brought about substantial advancements in the localization and classification of small objects, it is important to acknowledge the trade-offs involved. These include a large number of parameters (in the order of billions), several days of training (a few hundred epochs), and pretraining on extremely large datasets (which is not feasible without powerful computational resources). All of these aspects pose limitations on the pool of users who can train and test these techniques for their downstream tasks. It is now more important than ever to recognize the need for lightweight networks with efficient learning paradigms and architectures. Despite the number of parameters is now on par with the human brain, the performance in small object detection still lags considerably behind human capabilities, underscoring a significant gap in current research.

Furthermore, based on the findings presented in Figures \ref{COCO} and \ref{COCOzoom}, we have identified two primary challenges in small object detection: missing objects or false negatives, and redundant detected boxes. The issue of missing objects is likely attributable to the limited information embedded in the tokens. This can be resolved by using high-resolution images or by enhancing feature pyramids although this comes with the drawback of increased latency—which could potentially be offset by using more efficient, lightweight networks. The problem of repeated detections has traditionally been managed through post-processing techniques such as Non-Maximum Suppression (NMS). However, in the context of transformers, this issue should be approached by minimizing object query similarity in the decoder,  possibly through the use of auxiliary loss functions.

We also examined studies that employ transformer-based methods specifically dedicated to Small Object Detection (SOD) across a range of vision-based tasks. These include generic detection, detection in aerial images, abnormality detection in medical images, small hidden object detection in active millimeter-wave images for security purposes, underwater object detection, and small object detection in videos. Apart from generic and aerial image applications, transformers are underdeveloped in other applications, echoing observations made in  Rekavandi \textit{et al.} \cite{rekavandi2022guide} regarding maritime detection. This is particularly surprising given the potentially significant impact transformers could have in life-critical fields like medical imaging.

\section{Conclusion}\label{conclusion}
This survey paper reviewed over 60 research papers that focus on the development of transformers for the task of small object detection, including both purely transformer-based and hybrid techniques that integrate CNNs. These techniques have been examined from seven different perspectives: object representation, fast attention mechanisms for high-resolution or multi-scale feature maps, architecture and block modifications, spatio-temporal information, improved feature representation, auxiliary techniques, and fully transformer-based detection. Each of these categories includes several state-of-the-art (SOTA) techniques, each with its own set of advantages. We also compared these transformer-based approaches to CNN-based frameworks, discussing the similarities and differences between the two. Furthermore, for a range of vision applications, we introduced well-established datasets that serve as benchmarks for future research. Additionally, 12 datasets that have been used in SOD applications are discussed in detail, providing convenience for future research efforts.
In future research, the unique challenges associated with the detection of small objects in each application could be explored and addressed. Fields like medical imaging and underwater image analysis stand to gain significantly from the use of transformer models. Additionally, rather than increasing the complexity of transformers using larger models, alternative strategies could be explored to boost performance. 

\section{Acknowledgment}
We thank Likun Cai for providing the detection results for CBNet v2 in test images given in Figure \ref{COCO} and \ref{COCOzoom}. This research was partially supported by the Australian Research Council (ARC DP210101682, DP210102674) and Defence Science and Technology Group (DSTG) for the project “Low Observer Detection of Small Objects in Maritime Scenes”.
\ifCLASSOPTIONcaptionsoff
  \newpage
\fi
\bibliographystyle{IEEEtran}
\bibliography{egbib}

\end{document}